\documentclass[manuscript,screen]{acmart}
\pdfoutput=1

\AtBeginDocument{%
  \providecommand\BibTeX{{%
    \normalfont B\kern-0.5em{\scshape i\kern-0.25em b}\kern-0.8em\TeX}}}

\usepackage{arydshln}
\usepackage{algorithm}
\usepackage{algorithmic}
\usepackage{bbm}
\usepackage{float}
\usepackage{amsmath}
\usepackage{mathtools}
\usepackage{caption}
\usepackage{subcaption}
\usepackage{graphicx}
\usepackage{tikz}
\usetikzlibrary{positioning}

\usepackage{float}

\usepackage{setspace}
\begin{document}

\setcopyright{acmcopyright}
\copyrightyear{2023}
\acmYear{2023}
\acmDOI{XXXXXXX.XXXXXXX}

%%
%% These commands are for a JOURNAL article.
% \acmJournal{TOG}
% \acmVolume{37}
% \acmNumber{4}
% \acmArticle{111}
% \acmMonth{8}

\title{Learning Spatio-Temporal Aggregations for Large-Scale Capacity Expansion Problems}

\author{Aron Brenner}
\email{abrenner@mit.edu}
\affiliation{%
  \institution{Massachusetts Institute of Technology}
  \country{USA}
}

\author{Rahman Khorramfar}
\email{khorram@mit.edu}
\affiliation{%
  \institution{Massachusetts Institute of Technology}
  \country{USA}
}

\author{Saurabh Amin}
\email{amins@mit.edu}
\affiliation{%
  \institution{Massachusetts Institute of Technology}
  \country{USA}
}

% \renewcommand{\shortauthors}{Brenner, et al.}


\begin{abstract}
Effective investment planning decisions are crucial to ensure that critical cyber-physical infrastructures satisfy performance requirements over an extended time horizon. Computing these decisions often requires solving Capacity Expansion Problems (CEPs). In the context of regional-scale energy systems, these problems are prohibitively expensive to solve owing to large network sizes, heterogeneous node characteristics (e.g., electric power and natural gas energy vectors), and a large number of operational periods. To maintain tractability, traditional approaches resort to aggregating network nodes and/or selecting a set of representative time periods. Often, these reductions do not capture the supply-demand variations that crucially impact the CEP costs and constraints, leading to suboptimal decisions. 
Here, we propose a novel graph convolutional autoencoder approach for spatiotemporal aggregation of a generic CEP with heterogeneous nodes (CEPHN). Our autoencoder architecture leverages graph pooling to identify nodes with similar characteristics and minimizes a multi-objective loss function. This loss function is specifically tailored to induce desirable spatial and temporal aggregations in terms of tractability and optimality of CEPHN. In particular, the output of the graph pooling provides a \textit{spatial aggregation} while clustering the low-dimensional encoded representations yields a \textit{temporal aggregation}.
We apply our approach to generation expansion planning of coupled power and natural gas system in New England. The resulting spatiotemporal aggregation leads to a simpler CEPHN with 6 nodes (as opposed to 88 nodes in the original system) and a small set of representative days selected from a full year. We evaluate aggregation outcomes over a range of hyperparameters governing the loss function, and compare resulting upper bounds on the original problem with those obtained using previously known methods. The results from our case study show that this approach provides solutions that are 33\% (resp. 10\%) better those than obtained from standard spatial (resp. temporal) aggregation approaches. 

\end{abstract}

%     \centering
%   \includegraphics[width=0.8\textwidth]{diagram}
%   \caption{Spatio-temporal aggregation is often needed to tractably solve capacity expansion problems. Spatial aggregation is applied to reduce the number of decision variables and constraints associated with individual nodes while temporal aggregation is applied to reduce the number of decision variables and constraints associated with operational periods.}

\begin{CCSXML}
<ccs2012>
   <concept>
       <concept_id>10010147.10010257.10010258.10010260.10003697</concept_id>
       <concept_desc>Computing methodologies~Cluster analysis</concept_desc>
       <concept_significance>500</concept_significance>
       </concept>
   <concept>
       <concept_id>10010405.10010481.10010484.10011817</concept_id>
       <concept_desc>Applied computing~Multi-criterion optimization and decision-making</concept_desc>
       <concept_significance>300</concept_significance>
       </concept>
 </ccs2012>
\end{CCSXML}

\ccsdesc[500]{Computing methodologies~Cluster analysis}
\ccsdesc[300]{Applied computing~Multi-criterion optimization and decision-making}
% %%
% %% Keywords. The author(s) should pick words that accurately describe
% %% the work being presented. Separate the keywords with commas.
\keywords{Dimensionality Reduction, Clustering, Graph Neural Networks, Mixed Discrete/Continuous Optimization, Power System Planning}


\maketitle

\section{Introduction}
% \textcolor{red}{Revisions for ICCPS}
% \begin{itemize}
%     \item Change order of paper to have Optimization before GAMES
%     \item cite papers that use ML in decomposition methods. Highlight the fact that this paper's method can be used in conjunction with those. 
% \end{itemize}

Capacity Expansion Problems (CEPs) are a class of optimization problems that determine the optimal timing, location, and sizing of future expansion for a portfolio of assets to support the growing demand \citep{AhmedSahinidis2003approximation,GengJiang2009}. 
CEPs have found a broad range of applications in critical infrastructure networks, including communication systems \citep{RiisAndersen2004}, railway network \citep{RosellCodina2020},  urban water supply infrastructure \citep{FragaEtal2017}, and energy systems \cite{CEP,SinghEtal2009,KoltsaklisEtal2018,BennettEtal2021}. These applications usually involve making decisions for a set of network locations over a planning horizon. In many settings, expansion planning requires modeling the coupling between strategic level investment decisions with operational decisions, making CEPs a large-scale mathematical programs.
For instance, generation expansion problems (GEPs) in power systems involve computing long-term generation expansion decisions, while accounting for operational costs and constraints (e.g., dispatch, power flow, load control, and storage). Recently, the scope of these problems has been expanded to joint planning for multi-vector energy systems; in particular the coupling with natural gas (NG) and hydrogen \citep{KhorramfarEtal2022,BodalEtal2020}. Thus, an outstanding challenge is to determine the spatial and temporal resolution of CEPs that enable us to utilize modern optimization solvers to obtain high-quality solutions for realistic problems instances. Here, we develop machine learning (ML) based approach to address this challenge. Specifically, we build on recent advances in deep learning for graph-structured data and develop a spatio-temporal aggregation approach for a class of CEPs. 

% For instance, in power systems where CEPs are categorized as generation expansion problems (GEP), the long-term generation and network expansion decisions are usually accompanied by determining the operational decisions (e.g., dispatch, storage, flow); increasingly, these decisions are characterized by their coupling with other energy vectors such as natural gas (NG) and hydrogen \citep{KhorramfarEtal2022,BodalEtal2020}. Furthermore, the need for modeling with higher spatio-temporal resolutions and uncertainty considerations for resiliency purposes exacerbates the tractability of such problems.%The inclusion of these joint planning and operational considerations across a large network and over an extended time horizon (multiple years or decades) makes CEP formulations computationally intractable for realistic problem instances.

Spatial and/or temporal aggregation of these CEPs has been the subject of many recent studies that seek to make  problems analytically tractable for single-vector energy systems. Examples of temporal reduction strategies are rolling-horizon approach and representative period selection. In the first approach, the problem is solved iteratively over a manageable size of the planning horizon until all the planning periods are covered \citep{BodalEtal2022}. The latter approach is usually employed in power system CEPs where the planning problem is solved for a small number of appropriately chosen time periods and the corresponding weights \citep{Teichgraeber2022Survey}. On the other hand, spatial aggregation involves finding network structures such that the problem can be tractably solved over a subset of nodes or a reduced-sized network \citep{Oh2009,Shi2012}. However, prior work on aggregation methods for CEPs, mostly relies on ad hoc techniques rather than systematically exploring the space of spatial and temporal aggregations. For example, in GEP, the bus-level network is  aggregated into state or zone-level network \citep{KhorramfarEtal2022,LiEtal2022-EJOR}.

In CEPs with an underlying network structure, such as GEPs, an ideal parameter aggregation must yield a formulation that can be solved efficiently, while adequately capturing the parameters across spatial and temporal dimensions. Given that the planning decisions depend crucially on the network structure and planning horizon, the choice of aggregation can significantly impact the CEP's solution \cite{schyska, merrick}. Previous work has explored the use of ML methods to learn high-quality temporal aggregations \cite{TeichgraeberBrandt2022}. However, learning \textit{spatial} aggregation in conjunction with temporal aggregation is an unexplored problem, especially in the context of multivector energy systems.
%Motivated by the potential of ML models in learning data-driven representations as well as recent developments in deep learning for graph-structured data, here we use a 
Our graph convolutional autoencoder approach attempts to fill this gas by considering CEPs with underlying network structure and heterogeneous nodes (HN) that are characteristics of multivector energy systems (e.g., electricity power generation and consumption nodes or NG supply and demand nodes). Importantly, nodes may differ in terms of expansion cost parameters and operational constraints arising from technological differences and modeling resolution of individual subsystems. We refer to such planning problem as CEPHN. Our approach can be potentially applied to expansion planning for new cyber-physical systems, enabling refinement of classical approaches in facility location problems. 
%for \textit{spatio-temporal} aggregation of capacity expansion problems with an underlying network structure and heterogeneous nodes (CEPHN). CEPHN is a generalization of CEPs with an underlying network structure and allows problems to have nodes that operate on different time resolutions or input parameters. Examples of CEPHN can be found in cyber-physical systems \rt{cite}, multi-vector energy planning \citep{KhorramfarEtal2022}, and facility location problems.

The proposed graph convolutional autoencoder approach captures (i) spatial correlations between input parameters (ii) physical interdependencies between the heterogeneous nodes and (iii) heterogeneous granularity of data for two or more coupled networks for power and NG nodes as our test case. 
%We consider demand data for both systems, and consider CF data for solar and wind plants to reflect the supply pattern in the renewable-dominated future grid. 
The graph convolutional architecture accounts for the network interactions by training both within and across heterogeneous nodes autoencoder that minimizes a multi-objective loss function for node-specific input data. Moreover, we incorporate graph pooling to automatically learn spatial aggregations that group nodes exhibiting similar input patterns. The training procedure is based on the loss function that reconstruction loss and clustering objective. Our work provides a new direction to make CEPHNs tractable by utilizing the automatically learned spatio-temporal aggregations, leanding to high-quality planning decisions for  critical infrastructure networks.

  %demand as well as wind and solar CF data, which can be readily incorporated into high quality spatio-temporal aggregations. Furthermore, our approach to identifying spatio-temporal aggregations can also enable an accurate estimation of the trade-off between costs (both investment and operational) and joint emissions from power and NG systems.\footnote{We believe this capability can have a significant societal impact by lowering the barriers to investment in renewable energy resources and alleviating reliability concerns in a low-carbon energy system.}

% In particular, we apply our approach to address several computational challenges associated with a GEP resulting from a joint planning of electricity power and natural gas (NG) to meet the mid-century decarbonization targets while satisfying the projected demands. % jointly determine optimal investment decisions for electricity and natural gas systems under operational constraints and policy considerations. The formulation adheres to a reason, which is introduced in \cite{AE}, notably models two main interdependencies between power and NG systems. The first interdependency captures the increasing role of gas-fired power plants in the generation mix of electricity production \cite{eiaWebsite2021,HeEtal-survey2018}. The second interdependency reflects the \textit{joint} emission of CO$_2$ in both systems.
Specifically, we focus on aggregating and solving an otherwise intractable joint power-gas planning problem formulated as a GEP. The network model underlying this formulation is fairly detailed, capturing various power-gas operating constraints and decarbonization policy directives over a given planning horizon.
The problem has heterogeneous nodes, each with a different temporal resolution \citep{KhorramfarEtal2022}. We demonstrate that our autoencoder model is  well-suited to learning latent embeddings of the spatio-temporal patterns in this GEP and yield better solutions than previous techniques. 
The investment decisions of the GEP we consider here are over a planning horizon with yearly granularity, while the operational decisions (e.g., unit commitment, power production, load shedding, and energy storage) require hourly or sub-hourly resolution. In our case, the computational difficulty in solving the GEP increases further because we model both power and NG networks. Thus, considering the demand information at a day-to-day granularity becomes prohibitively expensive from a computational viewpoint. 
Prior work has tried to tackle such issues by aggregating power system nodes (buses) within a geographical neighborhood (power zone) to a single node~\citep{KhorramfarEtal2022,LiEtal2022-EJOR} and by solving the GEP for a set of representative periods (e.g., days) ~\citep{Teichgraeber2022Survey,HoffmannEtal2020-survey}. In contrast,  our approach accounts for the spatio-temporal variability in supply and demand patterns in determining the reduced dimensional network stricture and also provides the set of representative days. This yields a tractable CEPHN formulation that can be solved to obtain high-quality planning decisions

%formulation instantiated on this spatio-temporal aggregation of parameters should roughly capture the distribution of demand and supply patterns across the network and throughout the planning horizon.
 %To the best of our knowledge, the notion of representative days has not been clearly defined and developed in the context of joint power-NG planning problem. 
% Further, data-driven \textit{spatial} aggregations, in contrast to aggregations based on regional boundaries or more ad hoc specifications, have not been studied in the context of CEP models.

%% Our work also addresses \emph{practical issues} arising from sparse data availability from NG networks as compared to power system data. Firstly, while detailed connectivity and transmission information is readily available for many power networks, it is often the case that this data is not available to researchers for the corresponding NG network. Secondly, power systems typically collect demand and generation data at a fine temporal resolution (hourly or less), but this data is usually not publicly accessible for NG systems. These issues thus require us to (a) formulate network constraints based on loosely specified information on power and NG node connectivity and (b) develop an approach to leverage demand and supply data from the power system with demand data of NG system despite their different temporal resolutions.

Previous work on selecting sets of representative days using variants of k-means \cite{MallapragadaEtal2018,LiEtal2022,TeichgraeberBrandt2019,BarbarMallapragada2022},  k-medoids \cite{ScottEtal2019,TeichgraeberBrandt2019}, and hierarchical clustering \cite{LiuEtal2017,TeichgraeberBrandt2019}.
The distance data used in clustering algorithms are usually constructed based on a set of time series inputs such as load data and variable renewable energies (VRE) capacity factors \cite{LiEtal2022-EJOR,HoffmannEtal2020-survey}. Notably, these approaches neither account for demand data with multiple time resolutions nor spatial interdependencies, which we exploit with our graph convolutional approach. Hence, they cannot be readily extended to extract representative days for joint power-NG systems, an aspect that is crucial for realism and tractability in joint planning models for decarbonizing these systems. Spatial aggregation methods in power systems employ various techniques including bus elimination, node clustering based geographical boundaries, or other heuristic methods such as Ward reduction \citep{Shi2012,Oh2009}. However, these studies do not apply a fully data-driven approach to aggregating buses in the presence of another energy vector across nodes. We believe that our approach addresses these limitations.% and can provide a promising path to better extract spatio-temporal aggregations for other classes of CEPs with heterogeneous nodes.

 In the remainder of the paper we first present a generic CEP with heterogeneous nodes in Section  2. We define spatio-temporal aggregation and present relevant notation in Section 3. We review the autoencoder technique and give details of the graph convolutional neural network approach in Section 4. Section 5 presents the details of the case study, and Section 6 describes the setup for our numerical analysis. The results and discussion is provided in Section 7. Finally, we conclude the paper and provide future directions in Section 8.

\section{Capacity Expansion Model}\label{sec:cep}
In this section, we present a generic CEPHN for a network $\mathcal{G}=(\mathcal{N},\mathcal{E})$ with each node $n\in\mathcal{N}$ belonging to $s\in\mathcal{S}$, the set of node types. In the context of generation expansion models, different node types can represent nodes associated with each energy vector (e.g., power nodes, NG nodes, and storage nodes). We assume that the planning horizon for all nodes is the same, but the node types can differ in their time resolution required for operational planning. To incorporate heterogeneous times-scale of different node types, we assume that larger timer resolution are all multipliers of smallest one. For example, we can model power system operation on an hourly basis and NG network operations on a daily basis.   For a given class of nodes, $s$, we denote by $\mathcal{N}^s$ the set of nodes. Likewise, we denote by $\mathcal{T}^s$ the set of operational periods, e.g. hours or days. We introduce $\boldsymbol{x}^s$ as node-wise strategic-level decision variables, such as investment in or establishment of a new asset for class $s$. Similarly, we define $\boldsymbol{y}^s$ to be node-wise operational decisions for class $s$ over the set of planning periods $\mathcal{T}^s$. The CEPHN problem can then be formulated as:
% \begin{subequations}\label{model:cephn}
% \begin{align} 
% \min \ & \theta = (\CEone \xa + \CEtwo \ya) +(\CGone \xb + \CGtwo  \yb) \label{obj}\\
% \text{s.t.} \ & \Ae \xa+\Be\ya   \leq \bEone\label{e-c1}\\
%     %& \He \ye \geq \bEtwo& \label{e-c2}\\
%     & \Ag \xb + \Bg \yb  \leq \bGone \label{ng-c1}\\
%     % &  \fg = \Eone \pe %\texttt{ \bt{coupling constraint 1}}\text{[CC]}\qquad    &\label{coup1}\\
%     &\Ea\ya + \Fb \yb \leq \bab \label{coup2}
%     % &\xe \in \mathbb{Z}^+, \ye, \xg \in \mathbb{Z}^+\times \mathbb{R}^+, \pe, \yg,  \fg\in \mathbb{R}^+ & \label{e-c4}
% \end{align}
% \end{subequations}
\begin{subequations}\label{model:cephn}
\begin{align} 
\min_{\boldsymbol{x}^s,\boldsymbol{y}^s} \quad & \sum_{s \in \mathcal{S}} \sum_{n \in \mathcal{N}^s} \left(\mathbf{f}_n^s x_n^s + \sum_{t\in\mathcal{T}^s} \mathbf{g}_{nt}^s y_{nt}^s\right) \label{obj}\\
\text{s.t.} \quad & \mathbf{A}^s \boldsymbol{x}^s+\mathbf{B}^s \boldsymbol{y}^s \leq \mathbf{b}^s, \hspace{3mm} \forall s\in\mathcal{S}\label{e-c1}\\
    %& \He \ye \geq \bEtwo& \label{e-c2}\\
    % &  \fg = \Eone \pe %\texttt{ \bt{coupling constraint 1}}\text{[CC]}\qquad    &\label{coup1}\\
    &\sum_{s\in S}\mathbf{C}^s\boldsymbol{y}^s \leq \mathbf{h}^s \label{coup2}\\
    &\boldsymbol{x}^s \in \mathbb{Z}^{p^s}\times\mathbb{R}^{q^s}, \hspace{3mm} \forall s\in\mathcal{S}\\
    &\boldsymbol{y}^s \in \mathbb{R}^{l^s}, \hspace{3mm} \forall s\in\mathcal{S}
    % &\xe \in \mathbb{Z}^+, \ye, \xg \in \mathbb{Z}^+\times \mathbb{R}^+, \pe, \yg,  \fg\in \mathbb{R}^+ & \label{e-c4}
\end{align}
\end{subequations}

The objective function \eqref{obj} minimizes the total cost of investments and operations summed over the planning horizon. ~\eqref{e-c1} defines constraints in each class. The constraint~\eqref{coup2} couples operational constraints across all types of nodes. Without loss of generality, we assume that the shared variables only appear in the coupling constraint which is defined in terms of the parameters of that particular node type. In a multivector energy system, the coupling constraints model the exchange of energy between node types. They also capture policy constraints such as system-wide emissions reduction goals that limit the emission of CO$_2$ induced by all node types \citep{KhorramfarEtal2022, VonWaldEtal2022,BodalEtal2020}. In particular, our formulation can model a broad range  of CEPHNs from infrastructure planning \citep{GendreauEtal2006} to supply chain design \citep{MaEtal2020multi}. For most applications, this is a large-scale mixed integer programs and is not tractable even with the power of modern optimization solvers. Our spatio-temporal aggregation approach seeks to address this challenge.  %The original form of this formulation for many applications is a large-scale program which are not tractable for the current commercial solvers. The aggregation process seeks to alleviate this challenge. 

\section{Spatio-temporal Aggregation}\label{sec:agg}
Conceptually, the aggregation process is a set of mathematical operations designed to reduce the complexity of the problem (1) by reducing the network size and number of operational periods, and consequently, reducing the number of variables and/or constraints in the original formulation. An ideal spatio-temporal aggregation provides a reduced-form model whose solution and objective values are close to those of the original formulation. However, obtaining an exact solution for the original problem is, by assumption, intractable. As such, ``proxy'' methods are proposed in the literature in order to evaluate the quality of spatio-temporal aggregations. Examples include i) comparison of the solution of the aggregated problem to a lower bound of the original problem \citep{YokoyamaEtal2019}; ii)evaluating specific parameters (e.g., susceptance of transmission lines) \citep{Shi2012,Oh2009}; iii) comparing objective values for feasible solutions to the original problem obtained from the aggregated problem; \citep{MallapragadaEtal2018,TeichgraeberEtal2020}. 

The literature on spatial aggregation is relatively sparse and rather ad hoc. These include aggregating nodes by geographical boundaries \citep{KhorramfarEtal2022,LiEtal2022-EJOR}, and exploiting the underlying technical structure of the problem \citep{Oh2009,Shi2012}.
On the other hand, the literature on temporal aggregation for large-scale CEPs is more developed. Here, rolling-horizon and clustering of planning periods are common methods. Although rolling-horizon approach results in a feasible solution to the original problem, it does not capture the variation in demand and supply parameters over the entire planning horizon. Therefore, the absence of extreme days in some look-ahead periods can render an inferior solution, limiting the applicability of this method to realistic planning problems.
% A spatio-temporal aggregation method yields a problem which is defined over two heterogeneous nodes types $(\hat{\mathcal{N}}^{a}, \hat{\mathcal{N}}^{b})$ over $(\hat{\mathcal{T}}^{a}, \hat{\mathcal{T}}^{b})$ where 
% \begin{align*}
%     &|\hat{\mathcal{N}}^{a}| \leq |\mathcal{N}^a| \text{ and } |\hat{\mathcal{N}}^{b}| \leq |\mathcal{N}^b|\\
%     &
% |\hat{\mathcal{T}}^a| \leq |\mathcal{T}^a| \text{ and } |\hat{\mathcal{T}}^b| \leq |\mathcal{T}^b|
% \end{align*}
% Let $Y,\hat{Y}$ be the optimal solutions of the full problem and the aggregated problem. The goal of an aggregation methods is then to maximize $Sim (f(Y),f(\hat{Y}))$ where $Sim$ is a similarity measure of choice. The functions $f(Y)$ and $f(\hat{Y})$ can represent objective function value or any values associated with solution or parameters of these problems. 

\begin{figure*}
    \centering
    \includegraphics[width=0.85\textwidth]{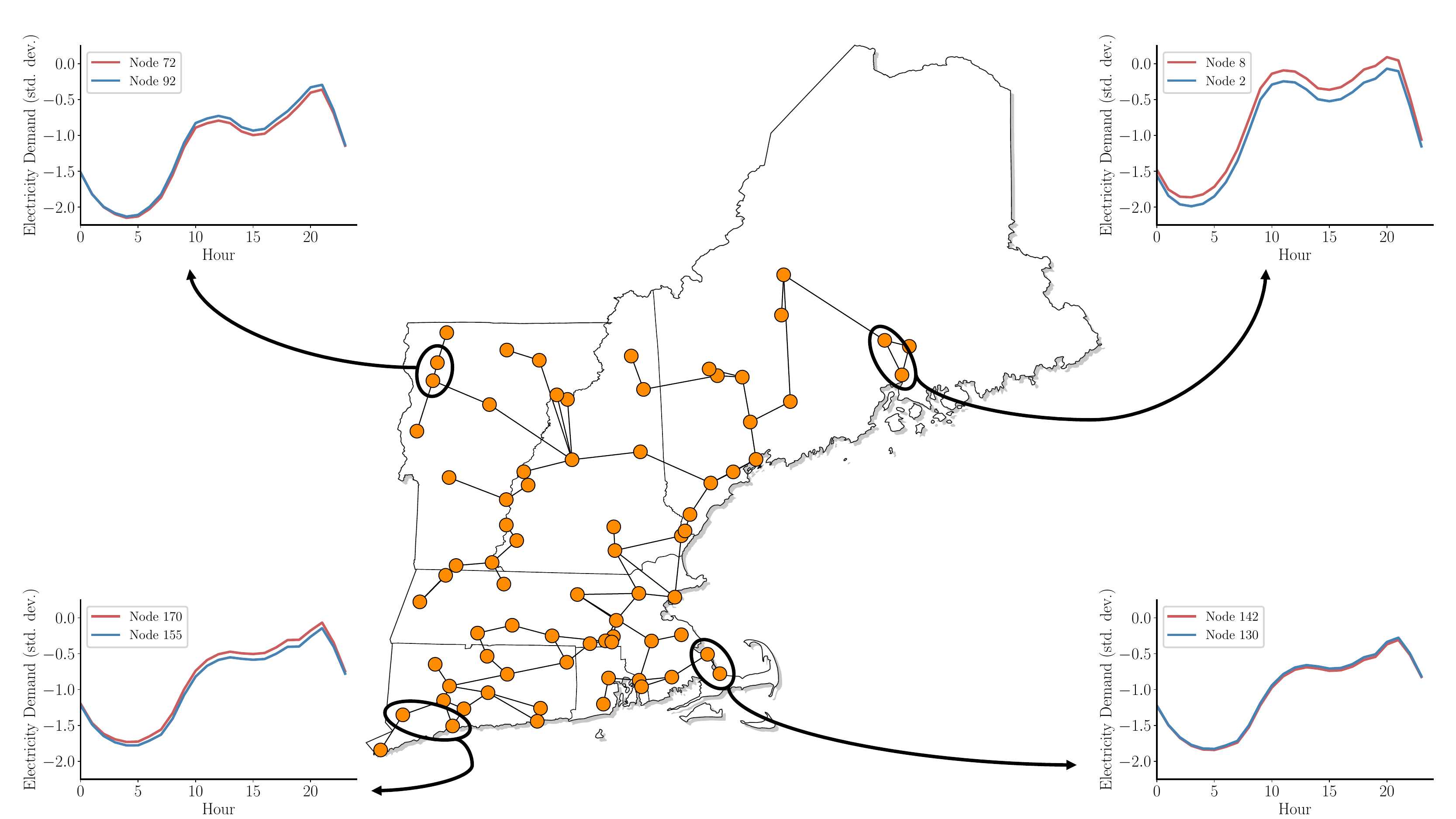}
    \caption{Demand nodes in the same neighborhood tend to demonstrate similar variations in electrical load over the course of the day. Spatial dependencies in CEPHN parameters, like those shown in this figure, are modeled explicitly by graph convolutional layers in the proposed autoencoder architectures.}
    \label{fig:diagram}
\end{figure*}

On the other hand, clustering methods aim to group the parameters associated with each operational period. For these methods, a distance metric, e.g., Euclidean distance, must be specified to measure dissimilarity between the parameters associated with any pair of operational periods. Variants of this method are employed in energy system planning problems, in particular k-means, k-medoids, and heirarchical clustering \citep{Teichgraeber2022Survey}. It is also common in power systems applications to construct a derived feature set by solving the full problem for a small number of planning periods and obtain different cost components or other output features for each planning period \citep{BarbarMallapragada2022,Teichgraeber2022Survey}. These derived features are then included in the distance calculation used by the clustering algorithm. In addition to the model parameters and derived features, clustering can also utilize other exogenous data that impact temporal variation in the parameters. In the power system context, such data can include ambient temperature, and electricity prices \citep{ZattiEtal2019}, which are known to have a significant impact on the realized power loads.

To build the case of our spatio temporal aggregation approach, we now illustrate clustering-based temporal aggregation on the generic CEPHN introduced in Section  2. The first step is to select a node type $s$ and its associated time resolution $t$. Recall that we assumed all time resolutions are multiplier of the smallest one. For each $s\in \mathcal{S}$, we construct the vector $\mathbf{x}_n^{(t)}$ for node $n$ at time $t$ as follows:
\begin{align*}
    \mathbf{x}_n^{(t)} = \left( 
    %\tilde{\mathbf{g}}_t \mathbin\Vert 
    %\tilde{\mathbf{P}}_t 
    %\mathbin\Vert \tilde{\mathbf{h}}_t 
    %\mathbin\Vert\mathrm{vec}(\tilde{\mathbf{A}}_t) 
    \mathrm{vec}(\tilde{\mathbf{P}}_n^{(t)}) %\mathbin\Vert\mathrm{vec}(\tilde{\mathbf{C}}_t) 
    \mathbin\Vert\mathrm{vec}(\tilde{\mathbf{O}}_n^{(t)}) 
    \mathbin\Vert\mathrm{vec}(\tilde{\mathbf{E}}_n^{(t)}) 
    \right),
\end{align*}
where $\mathbin\Vert$ and $\mathrm{vec}(\cdot)$ are the concatenation and vectorization operators, respectively. The elements of $\mathbf{x}_n^{(t)}$ include time varying information that captures the inherent variability across operational periods; the element  $\tilde{\mathbf{P}}^{(t)}_n$ denote a subset of CEPHN model parameters, $\tilde{\mathbf{O}}^{(t)}_n$ denote the aforementioned derived features, and $\tilde{\mathbf{E}}^{(t)}_n$ are exogenous data relevant for aggregation. We will refer to the elements of $\mathbf{x}_n^{(t)}$ henceforth as the observed \textit{features} of node $n$ at time $t$. We then construct a matrix of features for each subsystem $s \in \mathcal{S}$ by concatenating as follows
\begin{align*}
    \mathbf{X}_s^{(t)} = 
    \begin{bmatrix}
    \mathbf{x}_{s_1}^{(t)} \\
    \vdots \\
    \mathbf{x}_{s_{|\mathcal{N}^s|}}^{(t)} \end{bmatrix}
    \in\mathbb{R}^{|\mathcal{N}^s| \times d_s},
\end{align*}
%where $d_s$ is the total number of features and other data in $\mathbf{x}_n^{(t)}$.
where $\{s_1,\dots,s_{|\mathcal{N}^s|}\}$ is the set of nodes within subsystem $s$. In concatenation of features, if data of node type $s'\in \mathcal{S}\backslash s$ has coarser time resolution than $s$, then we repeat its features for each time period $t$. For example, suppose a CEPHN has two nodes types with hourly and daily resolution. If we choose to aggregate at an hourly temporal resolution, then data with a daily resolution will be repeated in all matrices, $\mathbf{X}_s^{(t)}$, for all one-hour periods $t$ within the same day. We also denote by $\mathcal{T}$ the set of time periods corresponding to the temporal resolution at which we aim to aggregate. After constructing the dataset $\{\mathbf{X}_s^{(t)} : s\in\mathcal{S}, t\in\mathcal{T}^s\}$ as described above, the parameters across all subsystems can be further vectorized and concatenated to construct
\begin{align*}
    \mathbf{x}^{(t)}=\left(\mathrm{vec}\big(\mathbf{X}_1^{(t)}\big) \mathbin\Vert \dots \mathbin\Vert \mathrm{vec}\big(\mathbf{x}_{|\mathcal{S}|}^{(t)}\big)\right),
\end{align*}
which has dimensionality $\sum_{s\in\mathcal{S}} |\mathcal{N}^s|d_s$. Then, a generic clustering algorithm such as k-medoids clustering can be applied to minimize the objective
\begin{align}
    \min_{\mathcal{C},\mathbf{z}} \sum_{k=1}^K \sum_{t\in \mathcal{C}_k} \|\mathbf{x}^{(t)} - \mathbf{z}_k\|\label{eq:kmedoids}
\end{align}
by assigning each of the parameter vectors to one of $K$ representative periods, where $\{\mathbf{z}_1,\dots,\mathbf{z}_K\} \subset \{\mathbf{x}^{(1)},\dots,\mathbf{x}^{|\mathcal{T}|}\}$ denote the cluster medians, $\mathcal{C}_k$ denotes the set of periods assigned to representative period $\mathbf{z}_k$, and $\|\cdot\|$ denotes a user-specified dissimilarity metric.

In practice, however, the performance of clustering algorithms deteriorate as the dimensionality of its input data increase \citep{BarbarMallapragada2022}.
This is due to overwhelming effect of random noise on distance calculations in high-dimensional settings, which poses a challenge to extracting meaningful distances between operational periods. \cite{BrennerEtal2022} and \cite{BarbarMallapragada2022} note that resolving this ``curse of dimensionality'' for aggregating large-scale CEPHNs, where large network sizes and high temporal resolutions of operational periods yield a very high-dimensional dataset.

\section{Learning Aggregation Architecture}
% We use a graph convolutional autoencoder approach to reduce the dimensionality of the graph-structured data. The autoencoder employs a pooling operation to minimize a multiobjective loss function. The pooling operation allows the autoencoder to identify nodes with similar patterns in their corresponding parameters (e.g., nodal demands, production capacities, etc.), hence facilitating spatial aggregation. The loss function, on the other hand, can be tuned to emphasize certain features of interest depending on the problem structure and importance of input data (see Section  \ref{sec:loss}). The output of the autoencoder is a low-dimensional and denoised representation of the data which is further used in a clustering algorithm to spatio-temporal aggregation \citep{parsons2004}.

To resolve the challenges associated with the high dimensionality of the dataset associated with the CEPHN problem, we develop a graph convolutional neural network (GCN) approach for spatio-temporal aggregation. Specifically, we train a graph convolutional autoencoder with a pooling operation (see Section  \ref{sec:GNN}) to minimize a multi-objective loss function, which can be tuned to trade off desirable properties of spatial and temporal aggregations (see Section  \ref{sec:loss}). By incorporating graph pooling, we constrain the autoencoder to identify groups of nodes that demonstrate similar patterns in their corresponding parameters (e.g., nodal demands, production capacities, etc.). This allows us to extract the learned pooling assignments as a \textit{spatially aggregated} network. Simultaneously, the pooled features can be extracted from the autoencoder, which yields low-dimensional representations of the data on which we can apply a clustering algorithm to obtain a \textit{temporal aggregation} \cite{parsons2004}. The resulting spatio-temporal aggregation yields a computationally tractable CEPHN formulation for a subset of operational periods and on a reduced network size.

% \begin{figure}[hbtp]
%     \centering
%     \includegraphics[width=\columnwidth]{aggregation_trade-off.png}
%     \caption{Increasing either the temporal resolution or the spatial resolution -- i.e. the number of representative days and nodes respectively -- will yield an instantiation closer to the original GEP. However, increasing both quickly yields an intractable formulation. Our objective is to understand the trade-off between aggregation mechanisms (e.g. spatial aggregation vs. temporal aggregation) with regard to faithfulness. In identifying a suitable aggregation, we must also aim for generalization. Indeed, aggregations may ``overfit'' to a dataset such that the aggregation that is most representative for the dataset might be less suitable for another dataset generated by the same process.}
%     \label{fig:trade-off}
% \end{figure}

\subsection{Overview of Autoencoders}
Autoencoders are deep neural networks that are trained to encode (compress) and decode (decompress) high-dimensional inputs. Unlike principal components analysis (PCA), autoencoders are non-linear models that are trained in a supervised fashion to minimize the error between input data and its corresponding reconstruction following compression \citep{GoodfellowEtal2016DeepLearning}. Importantly, these models learn to map high-dimensional inputs into lower-dimensional \textit{latent spaces}, and they are frequently used in practice to learn latent distances among high-dimensional data points while limiting the confounding effects of random noise.

Given a high-dimensional input such as a set of spatio-temporal CEPHN parameters, $\mathbf{x}^{(t)} \in \mathbb{R}^{d}$, an autencoder can be trained to jointly learn an encoder, $E_\phi: \mathbb{R}^d \to \mathbb{R}^{d'}$, and a decoder, $D_\theta: \mathbb{R}^{d'} \to \mathbb{R}^d$ that minimize the reconstruction loss function,
\begin{align*}
    \mathcal{L} = \sum_{t\in\mathcal{D}}\big\Vert\mathbf{x}^{(t)} - \hat{\mathbf{x}}^{(t)}\big\Vert_2^2
\end{align*}
over the dataset $\mathcal{D}$, where $\hat{\mathbf{x}}^{(t)}=D_\theta(E_\phi(\mathbf{x}^{(t)}))$ is the reconstructed input. Here, $d' \ll d$ denotes the dimension of the learned latent space, or more informally, the ``tightness'' of the compression bottleneck. Lower values of $d'$ will yield higher errors in reconstructing inputs, but are likely to be more effective in reducing the impact of noise on calculating distances to be used in clustering applications \cite{parsons2004}. Considering the temporal aggregation approach discussed in Section  \ref{sec:agg}, we have $d = \sum_{s\in\mathcal{S}} |\mathcal{N}^s|d_s$ and $\mathcal{D} = \mathcal{T}$. Then, one can instead aim to minimize (\ref{eq:kmedoids}) using distances computed from $\{E_\phi(\mathbf{x}^{(1)}),\dots,E_\phi(\mathbf{x}^{(|\mathcal{T}|)})\}$ rather than $\{\mathbf{x}^{(1)},\dots,\mathbf{x}^{(|\mathcal{T}|)}\}$.

Recent work on graph representation learning has facilitated the extension of deep unsupervised learning to high-dimensional graph-structured data \cite{vgae}. In Section  \ref{sec:GNN}, we introduce relevant ideas from modeling with graph neural networks. Then, in Section  \ref{sec:architecture} and \ref{sec:loss}, we describe our autoencoder architecture and training process.

\subsection{Graph Representation Learning}\label{sec:GNN}
To apply graph convolution operations, it is first necessary to encode a relevant graph structure over the nodes $\mathcal{N}$ with an affinity matrix, $\mathbf{A}\in\mathbb{R}^{|\mathcal{N}|\times |\mathcal{N}|}$. Importantly, the graph convolutions do not need to utilize the exact adjacencies, $\mathcal{E}$, given by the CEPHN network topology, $\mathcal{G}$. Rather, an appropriate affinity structure to encode is one that captures underlying interdependencies between nodes. For instance, one can choose to construct an affinity matrix based on geospatial distance, weather-related similarity, or demand patterns. We encode these affinities between any two node $n$ and $n'$ as
\begin{align*}
    \mathbf{A}_{nn'} = \exp{\left(-\frac{\mathrm{dist}(n,n')^2}{\sigma^2}\right)},
\end{align*}
where $\mathrm{dist}(n,n')$ denotes a distance metric of choice between nodes $n$ and $n'$ (e.g., the Euclidean distance between their corresponding geographical coordinates), and $\sigma$ denotes the standard deviation of distances in the network \cite{shuman2012}. We also construct the diagonal degree matrix $\mathbf{D}$ such that $\mathbf{D}_{nn} = \sum_{n'} \mathbf{A}_{nn'}$.

Following \cite{kipf2017}, we utilize \textit{Chebyshev convolutional filters}, which approximate spectral convolutions to learn latent node features as weighted local averages of observed and learned features for adjacent nodes. This is ideal for learning low-dimensional representations of networks as neighborhoods of nodes typically exhibit related (either similar or complementary) spatio-temporal patterns in demand, production capacity, etc. (see Fig. \ref{fig:diagram}). Chebyshev filters operate on the ``renormalized'' graph Laplacian $\tilde{\mathbf{L}} = \tilde{\mathbf{D}}^{-\frac{1}{2}}\tilde{\mathbf{A}}\tilde{\mathbf{D}}^{-\frac{1}{2}}$, where $\tilde{\mathbf{D}} = \mathbf{I} + \mathbf{D}$ and $\tilde{\mathbf{A}} = \mathbf{I} + \mathbf{A}$, and perform a form of Laplacian smoothing \cite{li2018, taubin1995}. We initialize $\mathbf{H}_{0} \in \mathbb{R}^{|\mathcal{N}|\times d}$ to be equal to our input matrix for $|\mathcal{N}|$ nodes with $d$ node features and apply convolutional filters to learn subsequent node features as follows:
\begin{align*}
    \mathbf{H}_{l+1} = \sigma (\tilde{\mathbf{L}} \mathbf{H}_{l} \boldsymbol{\Theta}_{l}),
\end{align*}
where $\boldsymbol{\Theta}_{l}$ is a trainable weight matrix and $\mathbf{H}_{l}$ is a matrix of node embeddings in layer $l$. $\sigma(\cdot)$ is typically a nonlinear activation function, such as $\mathrm{ReLU}$ or $\mathrm{tanh}$.

In each layer, GCNs aggregate features from the immediate neighborhood of each node. Deep GCNs stack multiple layers with nonlinear activations to learn latent node features as nonlinear functions of both local and global observed node features. In contrast, \cite{salha2019} propose a simpler graph autoencoder model, which they demonstrated to have competitive performances with multilayer GCNs on standard benchmark datasets despite being limited to linear first-order interactions. Shallow neural architectures are also better suited for settings where data availability is more limited, which is commonly the case for CEPHN applications. For example, CEPHN formulations for power systems often utilize thousands of parameters to represent daily power demand over the course of one day for a large network. However, the number of such observations may be on the order of several hundreds of days. Consequently, one must select the depth of the autoencoder keeping in mind this trade-off between generalization in highly limited data settings and capacity of the model to learn potentially richer node representations.

Driven by the need for dimensionality reduction and coarsening in graph-level ML tasks, multiple approaches for graph pooling have been developed in recent years \cite{diffpool,mincutpool,dmonpool}. A typical approach to graph pooling is to train a GCN block, or a series of graph convolutional layers, with a $\mathrm{softmax}$ output activation for node grouping assignments as part of a larger neural network. Given $|\mathcal{N}|$ nodes and $|\mathcal{N}'|<|\mathcal{N}|$ desired node groups, the output of this pooling assignment block is $\mathbf{S} \in \mathbb{R}^{|\mathcal{N}| \times |\mathcal{N}'|}$, a matrix of soft assignments in which $\mathbf{S}_{nn'} \in (0,1)$ encodes the degree to which node $n$ is assigned to group $n'$. To pool the node features after the $l$-th layer, $\mathbf{H}_{l}$, we simply compute $\mathbf{Z} = \mathbf{S}^\top\mathbf{H}_{l}$. Conversely, we can disaggregate the pooled feature matrix by computing $\mathbf{S}\mathbf{Z}$.

For example, one can imagine this block as taking in a time series of electricity demands for each node, $\mathbf{X}_s^{(t)}$, and returning a ``predicted'' membership distribution, parameterized by $\mathbf{S}^{(t)}$, which assigns all $|\mathcal{N}|$ nodes to one of the $|\mathcal{N}'|$ groups. Although there is no observed membership data with which to verify node group assignments, the quality of the assignments is evaluated implicitly by the reconstruction error $\mathcal{L}_R$ (which we define formally in Section  \ref{sec:loss}), and consequently the reconstruction loss is backpropagated to the pooling block during training of the autoencoder. To minimize reconstruction loss, the pooling mechanism will learn to pool nodes that exhibit similar electricity demands in each day.

For our application, we utilize the \textrm{MinCutPool} operator proposed by \cite{mincutpool}, which corresponds to including the objective
\begin{align}
    \mathcal{L}_P = \underbrace{-\frac{\mathrm{Tr}(\mathbf{S}^\top \tilde{\mathbf{A}}\mathbf{S})}{\mathrm{Tr}(\mathbf{S}^\top \tilde{\mathbf{D}}\mathbf{S})}}_{\mathcal{L}_C} + \underbrace{\left\|\frac{\mathbf{S}^\top\mathbf{S}}{\|\mathbf{S}^\top\mathbf{S}\|_F} - \frac{\mathbf{I}_{|\mathcal{N}'|}}{\sqrt{|\mathcal{N}'|}}\right\|_F}_{\mathcal{L}_O}.\label{eq:LP}
\end{align}
as part of a multi-objective loss function while training the autoencoder (see (\ref{eq:full_loss}) in Section  \ref{sec:loss}). This \textrm{MinCutPool} objective $\mathcal{L}_P$ is the sum of the \textit{cut loss}, $\mathcal{L}_C$, and the \textit{orthogonality loss}, $\mathcal{L}_O$. Minimizing the cut loss yields a pooling assignment that groups strongly connected nodes, while minimizing the orthogonality loss yields pooling assignments that are orthogonal and of similar sizes, i.e., each node is fully assigned to one group. The orthogonality loss term is included to discourage convergence to degenerate minima of the cut loss such as the uniform assignment of all nodes to all groups.

Moreover, it is often advantageous to ensure that node grouping assignments yield aggregated networks that are relatively uniform across certain parameters of interest. For example, one might prefer that the total demand of each node group (after summing over constituent nodes) does not differ too much from that of other groups. To this end, we also introduce the following \textit{negative entropy loss} objective
\begin{align}
    \mathcal{L}_H = (\mathbf{S}^\top \mathbf{H}_0 \mathbf{1})^\top \log (\mathbf{S}^\top \mathbf{H}_0 \mathbf{1}),\label{eq:LH}
\end{align}
where $\mathbf{1}\in\mathbb{R}^{d_s}$ is a vector of ones. Note that this negative entropy loss is minimized when the sums of input features for all node groups are identical. 

\subsection{Autoencoder Loss}\label{sec:loss}
Because CEPHNs consider different associated costs for investment and operational decisions across the different node classes (see Section  \ref{sec:cep}), we introduce hyperparameters $\{\alpha_1,\dots,\alpha_{|\mathcal{S}|}\}$ to tune the trade-off between the multiple reconstruction objectives, which gives us the total reconstruction loss objective
\begin{align}
    \mathcal{L}_R = \sum_{t\in\mathcal{T}} \sum_{s\in\mathcal{S}} \frac{\alpha_s}{|\mathcal{D}|} \big\Vert \mathbf{X}_s^{(t)} - \hat{\mathbf{X}}_s^{(t)}\big\Vert_F^2\label{eq:LR},
\end{align}
where $\|\cdot\|_F$ denotes the Frobenius norm. Then, the autoencoder can be trained to minimize the multi-objective loss function
\begin{align}
    \mathcal{L} = \alpha_R \mathcal{L}_R + \alpha_P \mathcal{L}_P + \alpha_H \mathcal{L}_H.\label{eq:full_loss}
\end{align}
The hyperparameters $\{\alpha_R, \alpha_P, \alpha_H\}$ weight the reconstruction, pooling, and entropy losses respectively. Altogether, we must specify up to $3+|\mathcal{S}|$ hyperparameters to weight the various objectives considered in this loss function, including those corresponding to reconstruction losses for different node classes. The class-specific reconstruction loss hyperparameters can be specified to reflect the contribution of each subsystem towards the total cost. Alternatively, we can choose to experiment with a range of these hyperparameters to determine the appropriate trade-offs between the multiple reconstruction losses and the pooling and entropy losses by solving the corresponding aggregated CEPHN and evaluating performance using its objective or an upper bound that it provides on the original (i.e., non-aggregated) CEPHN instance. In our case study, we evaluate a small range of these hyperparameter according to upper bounds on the original CEPHN given by solutions to aggregated instances, in which the aggregations are given by our autoencoder.

\subsection{GCN Architecture}\label{sec:architecture}
\begin{figure}[hbtp]
    \centering
    \includegraphics[width=0.7\textwidth]{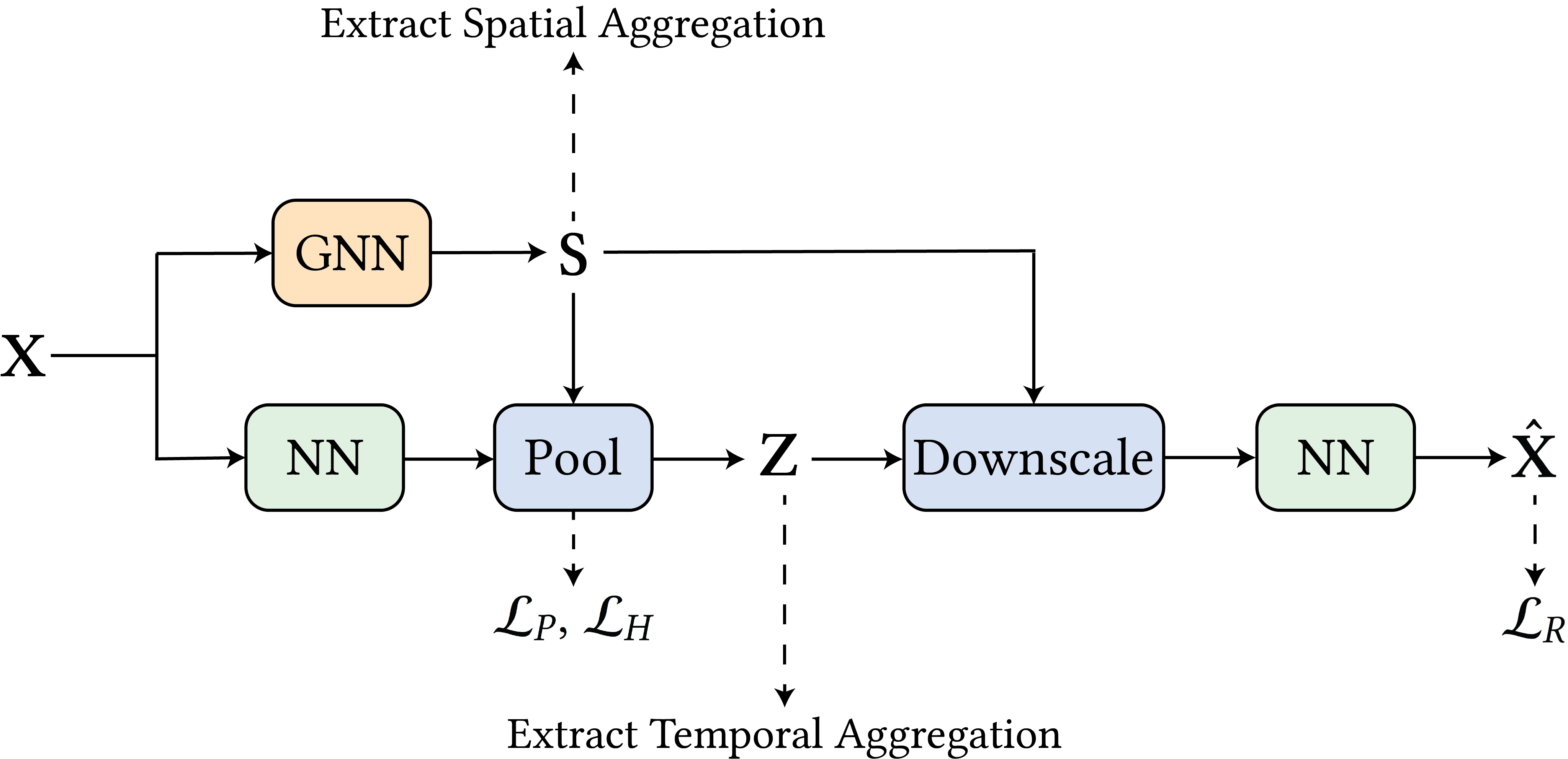}
    \caption{The autoencoder leverages a graph pooling operation to simultaneously learn spatial aggregations as node groupings and construct low-dimensional embeddings that can be used for identifying temporal aggregations via generic clustering algorithms. Values of component loss functions are extracted from different parts of the autoencoder, which must be balanced during the training process.}
    \label{fig:architecture}
\end{figure}

We now present the proposed GCN architecture for spatial and temporal aggregation, which is illustrated in Fig. \ref{fig:architecture}.
% \begin{table}[hbtp]
%     \centering
%     \begin{tabular}{|c|c|c|c|} 
%         \hline
%         \textbf{Variable} & \textbf{Interpretation} & \textbf{Granularity} & \textbf{Nodes} \\
%         \hline
%         $\mathbf{X}_E$ & Electricity & Hourly & 88 \\ 
%         \hline
%         $\mathbf{X}_W$ & Wind & Hourly & 88 \\
%         \hline
%         $\mathbf{X}_S$ & Solar & Hourly & 88 \\
%         \hline
%         $\mathbf{X}_G$ & Natural Gas & Daily & 18 \\
%         \hline
%     \end{tabular}
%     \caption{Notation for input variables.}
%     \label{tab:notation}
% \end{table}
To train the autoencoder, we first construct the input
\begin{align*}
    \mathbf{X}^{(t)} =
    \begin{bmatrix}
    \mathbf{X}_1^{(t)} & \mathbf{0} & \hdots & \mathbf{0} \\
    \mathbf{0} & \mathbf{X}_2^{(t)} & \hdots & \mathbf{0} \\
    \vdots & \vdots & \ddots & \vdots \\
    \mathbf{0} & \mathbf{0} & \hdots & \mathbf{X}_{|\mathcal{S}|}^{(t)}
    \end{bmatrix}
    \in\mathbb{R}^{|\mathcal{N}| \times (\sum_{s\in\mathcal{S}} d_s)},
\end{align*}
where $\mathbf{X}_s^{(t)} \in \mathbb{R}^{|\mathcal{N}^s|\times d_s}$ denotes the matrix of features corresponding to all nodes in class $s$ at time $t$. This block diagonal matrix effectively assigns a feature vector to each node with class-specific features for a subset of elements and zeros for all others. By applying graph convolutions, the autoencoder learns latent node features for a node in class $s$ using observed features from neighboring nodes across \textit{all classes} in $\mathcal{S}$. One may also include node-specific one-hot features in the input matrix, which is a common practice for graph convolutional modeling \cite{onehot}.

The encoder passes the input $\mathbf{X}^{(t)}$ in parallel through (1) a GCN block that outputs a node grouping assignment matrix, $\mathbf{S}^{(t)} \in (0,1)^{|\mathcal{N}|\times |\mathcal{N}'|}$, and (2) a second block that learns low-dimensional latent node features, $\mathbf{H}_{l}^{(t)}\in\mathbb{R}^{|\mathcal{N}| \times {d'}}$, using either fully connected layers or graph convolutions. Together, the outputs of these blocks are passed to a pooling layer, which aggregates the low-dimensional embeddings according to the learned grouping assignments, resulting in the low-dimensional node-aggregated embedding $\mathbf{Z}^{(t)} = \mathbf{S}^{(t)\top}\mathbf{H}_l^{(t)}$. This block also computes the $\mathrm{MinCutPool}$ loss, (\ref{eq:LP}) and the negative entropy loss, (\ref{eq:LH}). The pooled embedding matrix $\mathbf{Z}^{(t)}$ and node grouping assignment $\mathbf{S}^{(t)}$ are passed to the decoder.

To reconstruct the input, the decoder downscales the pooled feature matrix by computing $\mathbf{S}^{(t)}\mathbf{Z}^{(t)} \in \mathbb{R}^{|\mathcal{N}| \times {d'}}$. This matrix is then passed through several layers of graph convolutions and/or fully connected layers, which outputs the reconstructed parameter matrix $\hat{\mathbf{X}}^{(t)}$.

\subsection{Retrieving Spatio-Temporal Aggregations}
Once the autoencoder has been trained, each of the $|\mathcal{T}|$ operational periods can be passed through the encoder to retrieve the low-dimensional node-pooled feature matrices $\{\mathbf{Z}^{(1)},\dots,\mathbf{Z}^{(|\mathcal{T}|)}\}$. These feature matrices can then be vectorized into $({d'}|\mathcal{N}'|)$-dimensional vectors, and a generic clustering method, e.g. k-Medoids clustering, can be applied to identify a set of representative operational periods and their corresponding weights. Here, the clustering algorithm seeks to minimize the objective introduced in (\ref{eq:kmedoids}) using distances calculated from the $({d'}|\mathcal{N}'|)$-dimensional latent feature vectors $\{\mathbf{Z}^{(1)},\dots,\mathbf{Z}^{(|\mathcal{T}|)}\}$ rather than the $(\sum_{s}|\mathcal{N}^s|d_s)$-dimensional observed feature vectors.

Then, $|\mathcal{T}|$ spatial aggregations can be retrieved by passing each of the $|\mathcal{T}|$ observations through the pooling GCN block. From here, a single spatial aggregation can be chosen from the $|\mathcal{T}|$ candidates (many of which may be identical). One approach, which we chose for our case study, is to assign each node's respective group by taking a ``vote'' of group memberships over the $|\mathcal{T}|$ operational periods. Another possible approach would be to weight the vote based on the learned temporal aggregation (i.e., cluster median weights).

We note that the spatial and temporal aggregations do not need to be retrieved from the same trained autoencoder. Rather, one can select different inputs, hyperparameters, and architectures for spatial and temporal aggregation. Indeed, it is preferable to limit the inputs to only the most relevant parameters for aggregation as deep neural models can be trained more efficiently and with higher performance on lower dimensional inputs.

\section{Case Study}
We build on the joint power and natural gas planning problem proposed in \cite{KhorramfarEtal2022}. Although the authors introduce a generation and transmission expansion problem, here we consider a simpler problem that can be treated as a GEP; see Supplementary Information (SI) \cite{suppMat} for full description of the problem. The problem determines the minimum investment and operational costs of co-optimizing electricity and NG systems for the year 2050 under various investment, operational, and policy constraints. The investment decisions for the power system include establishing new plants and decommissioning existing plants while the decisions for the natural gas network include establishing new pipelines. Our model accounts for major planning constraints such as minimum stable production, ramping, energy balance, and storage. The interdependency between the two systems is captured by two sets of constraints. The first coupling constraint captures the flow of natural gas to the power system to enable the operation of gas-fired power plants. The second constraint imposes a system-wide emission constraint, limiting the emissions of CO$_2$ from both systems to a pre-specified value.

This GEP considers both electric power and NG nodes, each having their own input parameters and operating on different time resolutions; hence the problem is a CEPHN. The power system operates on an hourly basis whereas the NG system operations are carried out at a daily resolution. In the formulation presented in the SI, the sets of node types consist of power and NG (i.e., $\mathcal{S}=\{1,2\}$). The investment decisions for the power system, $\boldsymbol{x}^1$, include investment in the generation and storage technologies at each node. Some of the investment decisions, such as investment in thermal generation, are assumed to be integer decisions, while other variables are defined as continuous. The operational decisions for the power system include power generation, storage, and load shedding decisions, which are collected in $\boldsymbol{y}^1$. The variable $\boldsymbol{x}^2$ for the NG system includes investment in pipeline expansion, which is a binary variable. The operational variables $\boldsymbol{y}^2$ include fuel supply, flow, and load shedding. %In SI, the constraints (2a)-(6a) correspond to the constraints specific to power system, and constraint (8a)-(10c) are NG constraints. These constraints form the constraints (1b) in model (1). The coupling constraints (11a) in SI ensures that gas-fired plants operate based on the gas they receive from the NG network. The second coupling constraint (11b) in SI imposes the economy-wide decarbonization constraint. Together these constraints can be captured by constraint (1c) in model (1). 

We utilize the input data provided in \cite{KhorramfarEtal2022}. There, the authors construct a network model of the New England power and NG networks from publicly available data. The power network consists of 188 buses located in 88 distinct nodes (i.e., locations). The NG network consists of 18 NG nodes. The input data considers 12 power plants types including 5 existing and 7 new plant types. The plant types represent a range of generation options such as various gas-fired plants, wind, solar, hydro and nuclear power generators.% The SI \cite{suppMat} provides the details of the input data for the joint power-NG planning model \cite{suppMat}.

\section{Experimental Setup}
We use the case study in Section  5 to evaluate the performance of our approach and consider power demand, NG demand, and capacity factor (CF) as input data. For a given number of representative days, we obtain the following five aggregations for the GEP:
\begin{itemize}
    \item Aggregation based on U.S. states; nodes that are located in the same state are represented as a single node in the aggregated GEP. This aggregation is common practice in the literature and serves as a benchmark for our case study.
    \item Aggregation based on pooling loss only (PL); values of $\alpha_R, \alpha_P$, and $\alpha_H$ in Eq. (\ref{eq:full_loss}) are set to 0, 1, and 0 respectively.
    \item Aggregation based on pooling and reconstruction losses (PRL); values of $\alpha_R, \alpha_P$, and $\alpha_H$ in Eq. (\ref{eq:full_loss}) are set to 1, 1, and 0 respectively. 
    \item Aggregation based on pooling and entropy losses (PHL);  values of $\alpha_R, \alpha_P$, and $\alpha_H$  in Eq. (\ref{eq:full_loss}) are set to 0, 1, and 1 respectively. 
    \item Aggregation based on pooling, reconstruction, and entropy losses (PRHL); values of $\alpha_R, \alpha_P$, and $\alpha_H$  in Eq. (\ref{eq:full_loss}) are all set to 1.  
\end{itemize}

For temporal aggregation, we consider 5, 10, 20, and 40 representative operational periods out of 365 days. We obtain four values for each number of representative days. We use k-medoid to cluster the raw data which, includes demands and CFs. We also applied PCA to the raw data and clustered the resulting encodings. These two methods are commonly used in the literature. We also consider two outcomes from the proposed approach. In the first outcome (A1), we only consider power demand, whereas the second outcome (A2) is a result of considering power demand, NG demand, and CFs. This experimental setup results in 80 instances in total.

All instances are implemented in Python using Gurobi 10.0 and are available at the GitHub page \citep{suppMat}. All instances are run on the MIT Supercloud system, which uses an Intel Xeon Platinum 8260 processor with up to 48 cores
and 192 GB of RAM \cite{Supercloud2018}. We limit the CPU time to 3 hours for all instances, by which point all models were solved to optimality with a mixed integer gap of 1\% or lower.

\section{Results and Discussion}

To evaluate each instance of the spatio-temporal aggregation, we propose a heuristic approach that generates a feasible solution for the full problem (88 nodes and the entire year). The approach consists of three steps.
\begin{enumerate}
    \item We first solve the spatially and temporally aggregated problem and retrieve investment decisions for each cluster of nodes.
    \item We then consider the full network for only two representative days and add constraints to ensure that the sum of investment decisions over all nodes represented by a group do not exceed the number of investment decisions for the group in the first step. For example, suppose that five solar plants are established in the first step for node group 7, and assume that nodes 3, 6, and 44 are represented by group 7. Then in the second step, we constrain the sum of established solar plants across nodes 3, 6, and 44 to be at most five. The result of this step provides investment decisions for each node.
    \item Finally, we consider the full network over the whole year and fix the investment decisions returned by Step 2. This renders the problem as a linear program in which the only decision variables correspond to continuous-valued operational decisions.
\end{enumerate}
The solution to the third step provides a feasible solution to the original CEPHN, and consequently is an upper bound (UB).

\begin{figure}[hbtp]
     \centering
     \begin{subfigure}{0.45\textwidth}
         \centering
         \includegraphics[width=\textwidth]{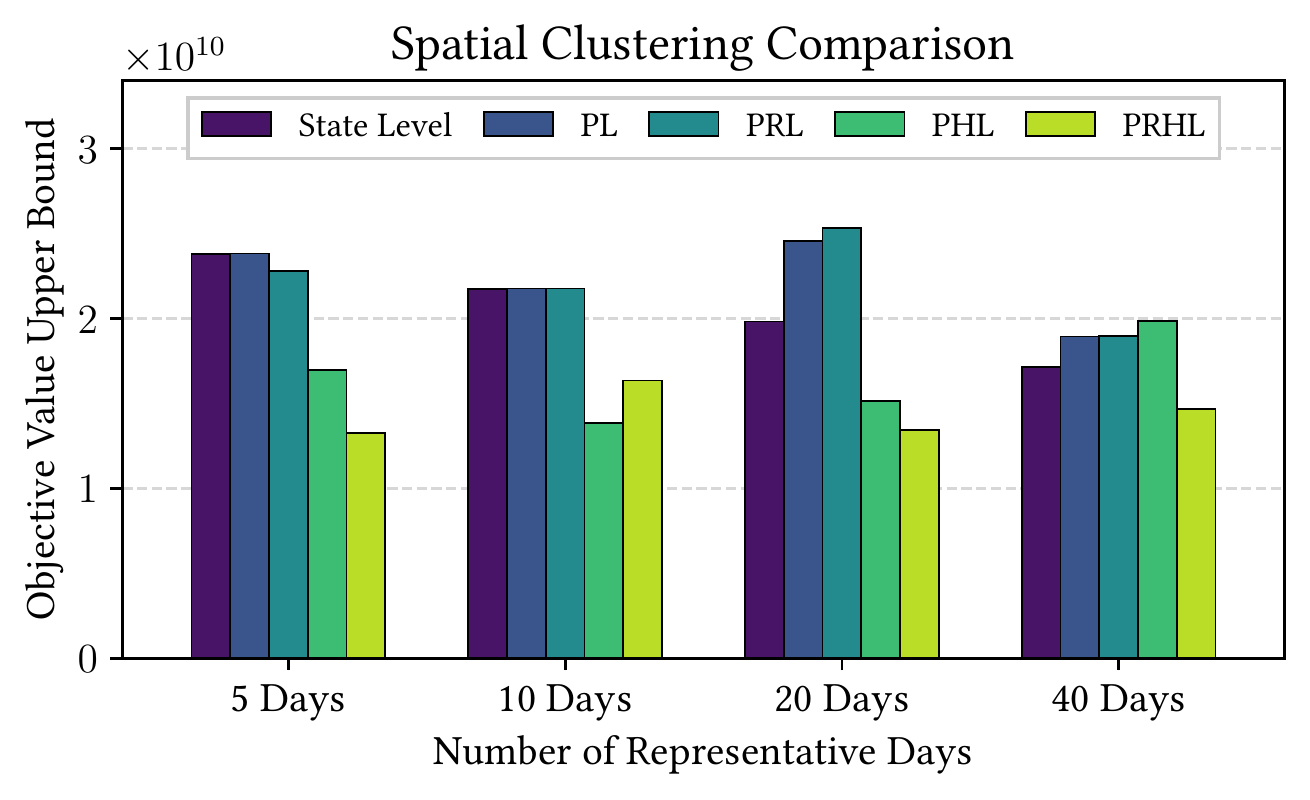}
         \caption{UB comparison for spatial aggregation methods. Each bar corresponds to the \textit{lowest} UB computed for the given spatial aggregation method across \textit{all} temporal aggregation methods.}
         \label{fig:spatial-res}
     \end{subfigure}
     \hfill
     \begin{subfigure}{0.45\textwidth}
        \centering
        \includegraphics[width=\textwidth]{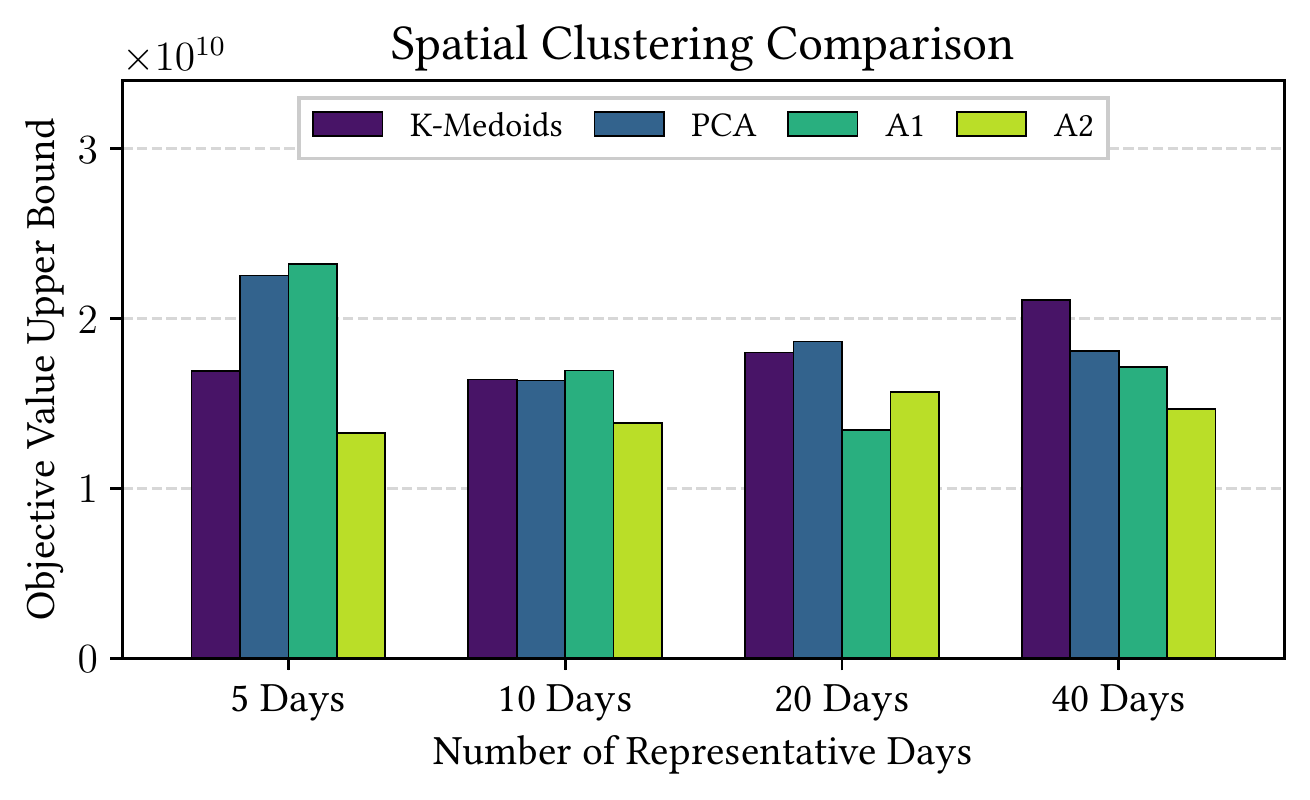}
        \caption{UB comparison for temporal aggregation methods. Each bar corresponds to the \textit{lowest} UB computed for the given temporal aggregation method across \textit{all} spatial aggregation methods.}
        \label{fig:temporal-res}
    \end{subfigure}
    %\caption{Upper Bound Comparison}
\end{figure}

Fig.~\ref{fig:spatial-res} illustrates the best performance of different spatial reduction methods across four temporal aggregation methods. We note that spatial aggregations from the proposed approach outperform in all representative days sizes. PRHL, which is trained to minimize the pooling, reconstruction, and negative entropy losses, outperforms all other methods including the state level aggregation commonly used in the literature for all but 10 representative days. Its performance is followed by PHL, which is trained to minimize the pooling and negative entropy losses. Overall, the upper bound obtained by instances that are spatially aggregated using PRHL is 33\%, 40\%, 39\%, and 10\% better than the upper bounds obtained from aggregating using state level, PL, PRL, and PHL, respectively (after averaging across all temporal aggregation methods and numbers of representative days). The result suggest that the inclusion of the negative entropy loss in conjunction with the reconstruction loss can provide superior spatial aggregations as evaluated by our 3-step upper bound procedure.

The best performance observed for the four temporal aggregation methods across the considered spatial aggregations is shown in Fig.~\ref{fig:temporal-res}.  A2, which is trained to minimize reconstruction error for power demands, NG demands, and CFs, outperforms the other temporal aggregation methods for all but 20 representative days. The feasible solution corresponding to the temporal aggregation from A1 yields the worst upper bound for 5 representative days, but A2 performs significantly better for the same number of representative days, highlighting the significance of including NG demand and CF data in training the autoencoder. Furthermore, A2 demonstrates more consistent upper bounds across different number of representative days. This is significant, as some large-scale CEPHNs can only be solved for a small number of representative days. Consequently, a temporal aggregation method that poorly captures the variability of supply-demand can have significant implications for planning and operations \citep{BennettEtal2021}. Overall, the investment decisions retrieved from solving the temporal aggregation using A2 yields an upper bound that is 10\%, 9\%, and 7\% better than k-medoid, PCA and A1, respectively (after averaging across all spatial aggregation methods and numbers of representative days).

\begin{figure*}
\centering
\begin{subfigure}{0.49\textwidth}
    \includegraphics[width=\textwidth]{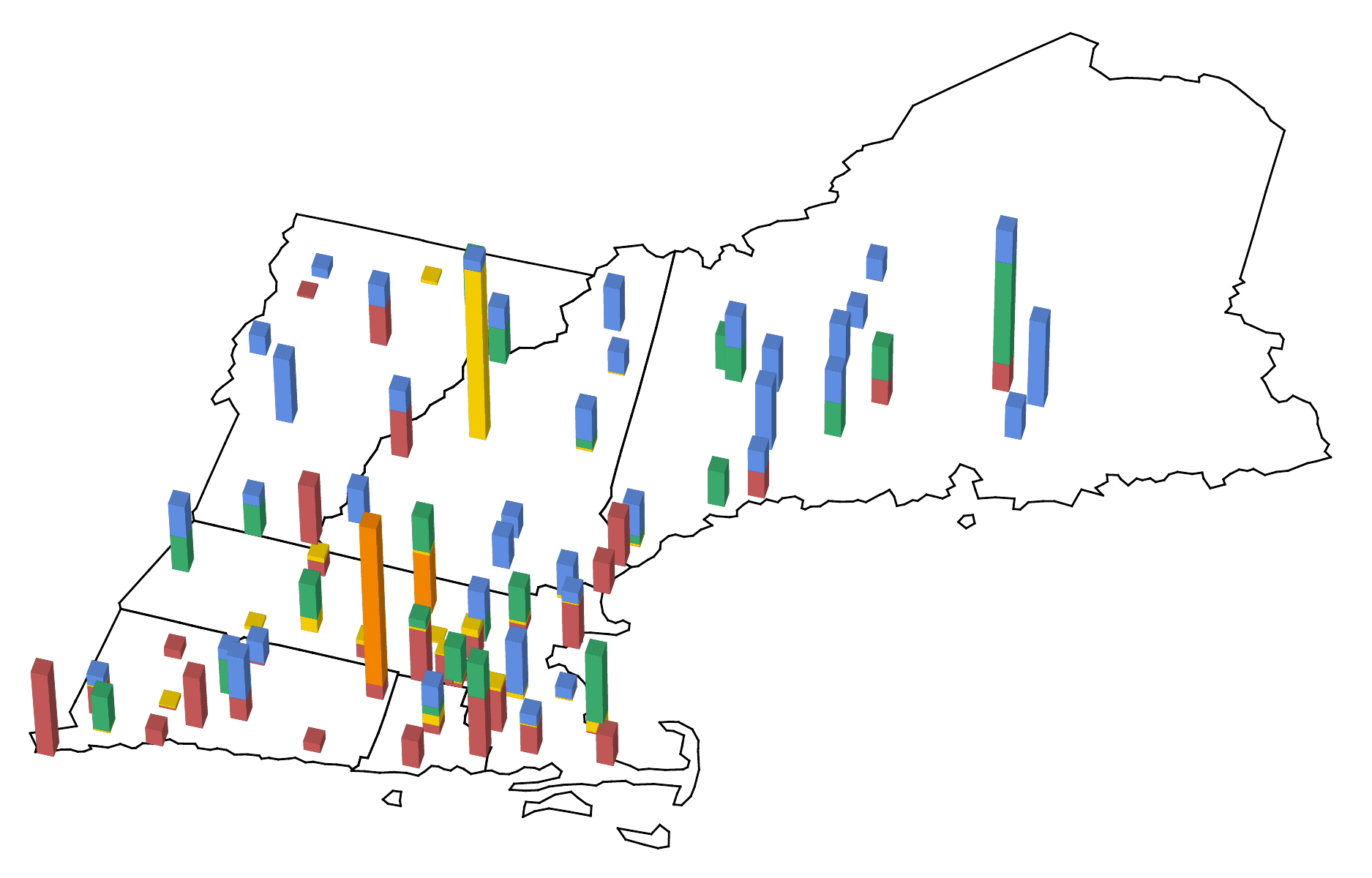}
    \caption{State-Level, A1, 10 Rep. Days}
\end{subfigure}
\hfill
\begin{subfigure}{0.49\textwidth}
    \includegraphics[width=\textwidth]{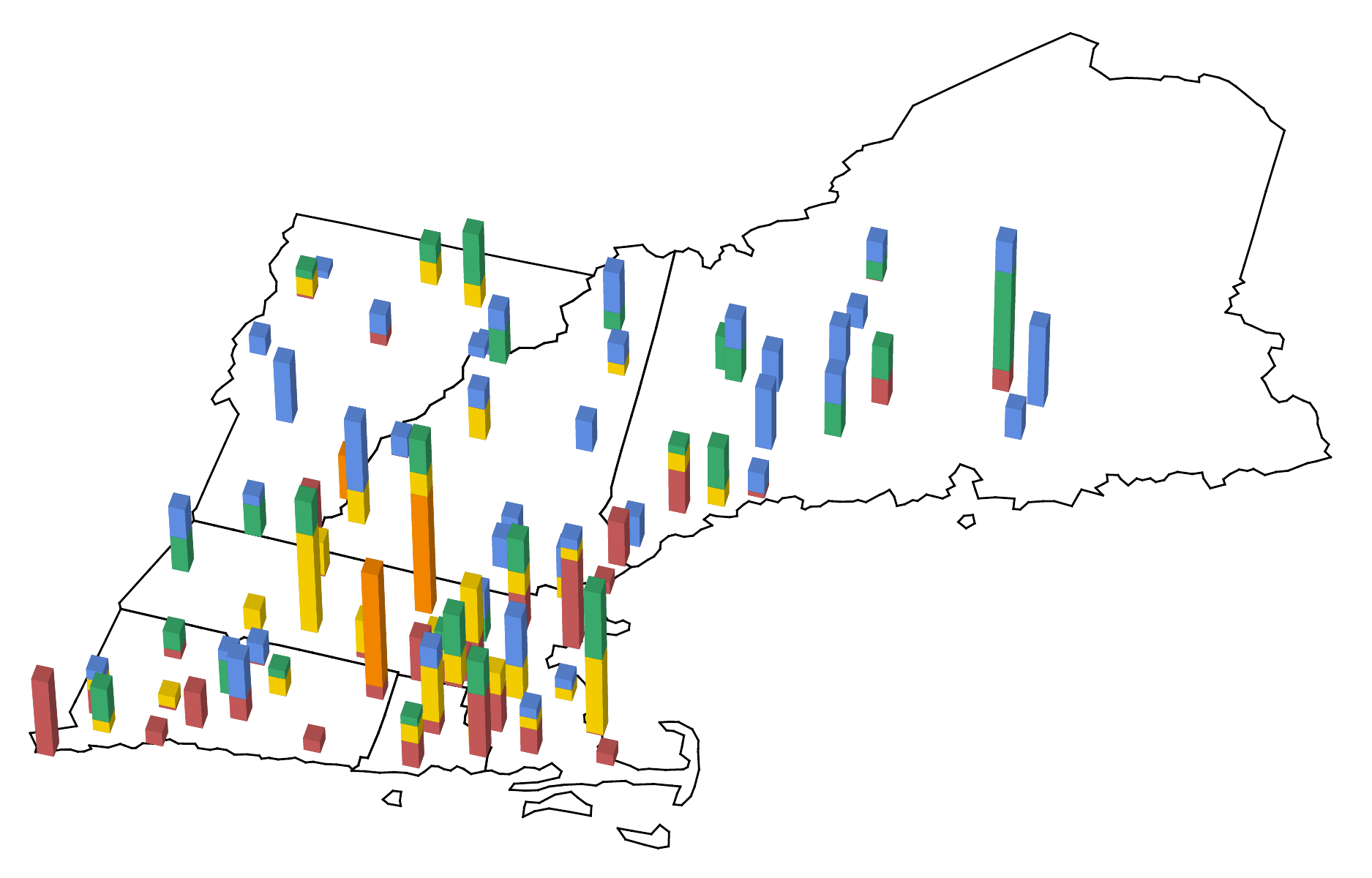}
    \caption{State-Level, A2, 10 Rep. Days}
\end{subfigure}
\hfill
\begin{subfigure}{0.49\textwidth}
    \includegraphics[width=\textwidth]{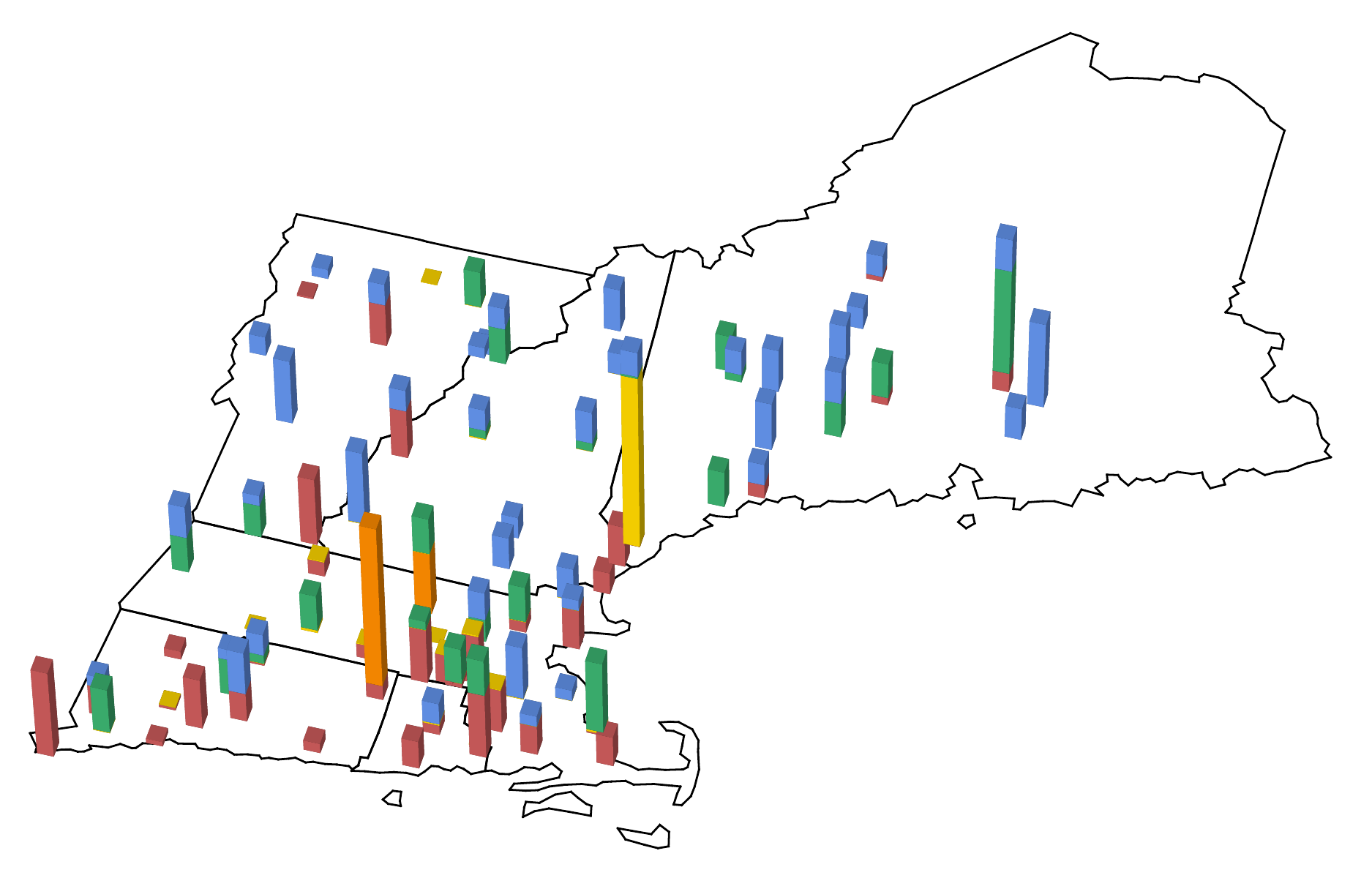}
    \caption{PRHL, A1, 10 Rep. Days}
\end{subfigure}
\hfill
\begin{subfigure}{0.49\textwidth}
    \includegraphics[width=\textwidth]{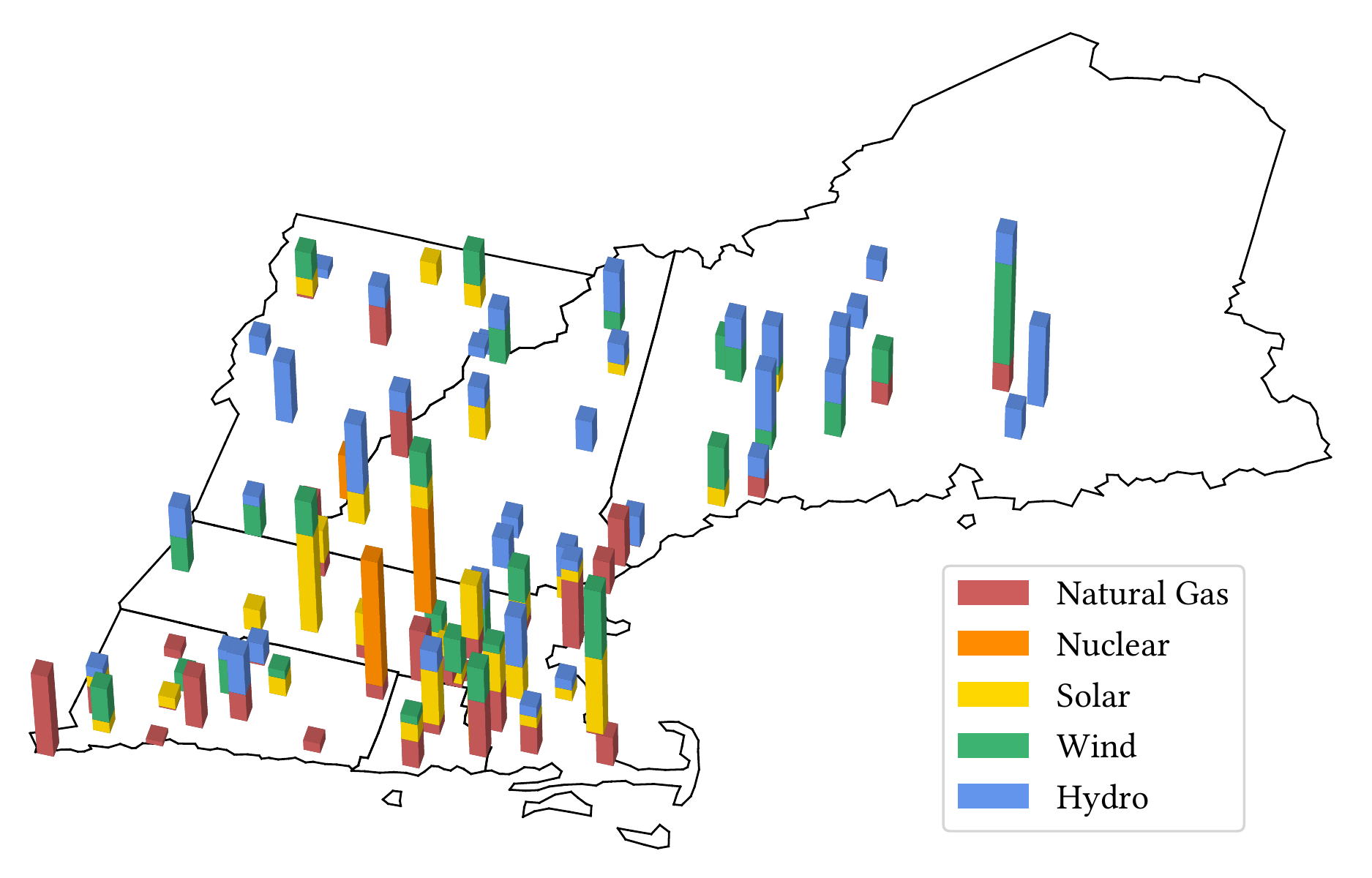}
    \caption{PRHL, A2, 10 Rep. Days}
\end{subfigure}
\caption{Generation mix comparison across four spatio-temporal aggregations. To make comparison easier, the height of each bar is normalized according to the total generation for its corresponding source.}
\label{fig:gen-mix}
\end{figure*}

Generation mix is another result that significantly impacts planning outcomes. Fig.~\ref{fig:gen-mix} illustrates power generation across different sources as solved for using two spatial and two temporal aggregation methods for 10 representative days. We compare state-level and PRHL with A1 and A2 to compare the generation outcomes from the proposed approach to aggregations commonly used in the literature. We find that the choice of temporal aggregation has a greater impact than the method for spatial aggregation. 
%, an observation also pointed by other studies \rt{cite}. 
In particular, the inclusion of NG demand and CFs in A2 significantly increases the share of solar generation and distributes its generation more uniformly across different nodes, whereas in A1, we observe solar generation is concentrated in a smaller set of locations. We also observe, albeit to a smaller degree, differences in generation from wind and NG resulting from the two temporal aggregation methods.

\section{Conclusion}
In this paper, we present a graph convolutional autoencoder approach to spatio-temporally aggregate capacity expansion problems with heterogeneous nodes. We leverage a graph pooling operation to automatically identify nodes exhibiting similar characteristics. Moreover, our proposed architecture utilizes graph convolutions to exploit network interactions within and across subsystems. We also introduce a multi-objective training loss which can be tuned to capture varying subsystem relevance and desirable qualities of spatial aggregations.

We apply our approach to a generation expansion problem for the New England joint power-NG network. Specifically, we experiment with reducing the original problem size from 88 to 6 nodes using various trade-offs between components of the multi-objective loss function. We also experiment with applying our temporal aggregation approach to retrieve two sets of representative days, one of which is obtained using only power demand data, while the other incorporates power demands, NG demands, and capacity factors for variables renewable generators (i.e., solar and wind). We consider common spatial and temporal aggregation methods from the literature and compare the solutions associated with each of these methods. We find that solutions resulting from our approach significantly outperform those obtained from existing methods for our case study.

% present two autoencoder architectures: an architecture for temporal aggregation that utilizes electricity demand, NG demand, wind CF, and solar CF data, and an architecture for spatial aggregation that uses electricity demand data to learn a node cluster assignment via graph pooling. Finally, we interpret the sensitivity of various cost components with respect to spatial and temporal aggregation resolution given a set of aggregations learned by our approach. Our analysis highlights the importance of data-driven aggregation methods and show that some highly aggregated model could provide quality solutions for capacity expansion problems.

Future studies may extend the approach and compute additional experiments made in this work in several directions. 
For one, additional experiments can be performed with different CEPHN formulations to more comprehensively understand the relationship connecting aggregation resolution to both solution quality and sensitivity of the learned planning decisions. Moreover, the sensitivity of aggregations, and consequently planning decisions, with respect to changes in the dataset should be investigated.

\section*{Acknowledgments}
This work is supported by funding from MIT Energy Initiative Future Energy Systems Center, MIT Climate Grand Challenges grant for ``Preparing for a new world of weather and climate extremes'', MIT School of Engineering Stockham Fellowship, and C3.ai grant for ``CLAIRE: Causal reasoning for real-time attack localization in cyber-physical systems''.

\bibliographystyle{ACM-Reference-Format.bst}
\bibliography{bib}

\section*{Appendix}
\subsection*{Model Formulation}
Our formulation is based on the formulation proposed in \cite{KhorramfarEtal2022}, but we applied a set of assumption to simplify the model and speed up the running time.  The model determines minimum cost investment and operational decision for power and NG system.
across a set of representative periods. The formulation allows different temporal resolutions for the operation of both systems as data availability or planning requirements can be different for power and NG systems. The operations of both systems are coupled through two sets of constraints. The first set ensure NG flow to the power system. The second coupling constraints limit the CO$_2$ emission incurred by consuming NG in both power and NG systems. 

\begin{figure}[htbp]
    \centering
    \includegraphics[width=0.4\textwidth]{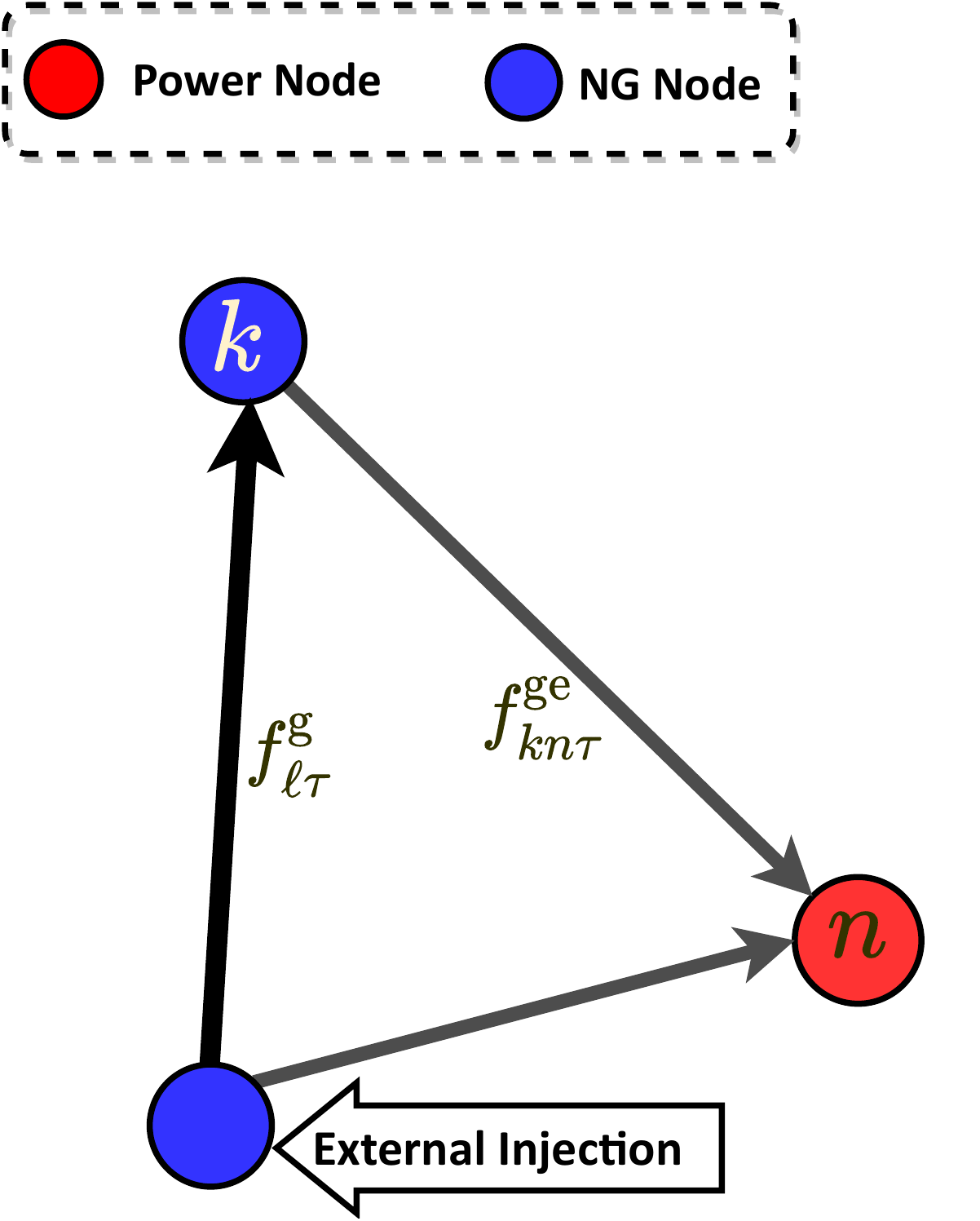}
    \caption{ Each power operate local NG-fired plants by drawing gas from nodes that are connected to it. The variable $f^{\text{ge}}_{\ell \tau}$ captures this flow. %the flow from NG node $k$ to power node $n$ at time period$\tau$. 
    Each NG node is connected to its adjacent SVL nodes through two unidirectional pipelines where one is from NG to SVL's liquefaction facilities denoted by $f^{\text{gl}}_{kj \tau}$; and the other one from SVL's vaporization facility to NG node denoted by $f^{\text{vg}}_{jk\tau}$. The variable $f^{\text{g}}_{\ell \tau}$ denotes the flow between NG nodes. NG nodes can be connected by one or more uni-directional pipelines, but only one connection is depicted here. Candidate pipelines are not shown in this figure.}
    \label{fig:var_nodes}
\end{figure}

The network consists of three sets of nodes as depicted in Figure~\ref{fig:var_nodes}. The first set represents power system nodes and is characterized by different generation technologies (plant types), demand, storage, and the set of adjacent nodes.
%NG system consists of NG and SVL nodes. 
The second set of nodes are NG nodes each of which associated with injection amount, demand, and its adjacent nodes. Storage tanks, vaporization and liquefaction facilities, which  are commonly used in the non-reservoir storage of NG, collectively form the third set of nodes namely SVL nodes. The model also consider the \textit{renewable natural gas} (RNG) which is type of net-zero biofuel fully interchangeable with NG and hence can be imported and transported by the NG pipelines \cite{ColeEtal2021}. Details regarding input data including generation plant and storage types, demand, and cost assumptions are provided in the Appendix of Supplementary Information of \cite{KhorramfarEtal2022}.%~\ref{App:input-data}. 

%\subsection{Notation}

\begin{table}[htbp]
\small
    \centering
    %\caption{Indices}
    \begin{tabular}{l l}
    \multicolumn{2}{l}{{Indices}}   \\
    \midrule 
    $n,m$ & Power system node \\
    $k$ & NG system node\\
    $j$ & SVL facility node\\ 
    $i$ & Power generation plant type\\
    $r$ & Storage type for power network\\
    $\ell$& Pipeline\\
    $t$ & Time step for power system planning \\
    $\tau$ & Time step for NG system planning \\
\bottomrule
    \end{tabular}
    \end{table}
    
\begin{table}
\small
    \centering
    %\caption{Sets}
    \begin{tabular}{l l}
      \multicolumn{2}{l}{{Sets}}\\
      \midrule
    $\mathcal{N}^{\text{e}}$ & Power system nodes\\
    $\mathcal{P}$ &  Power plant types 
    \\
    $\mathcal{R} \subset \mathcal{P}$ & VRE power plant types 
    \\
    $\mathcal{G} \subset \mathcal{P}$ & gas-fired plant types 
    \\
    $\mathcal{H} \subset \mathcal{P}$ & Thermal plant types  
    %(i.e. gas-fired plants and nuclear)
    \\
    % $\mathcal{Q} \subset \mathcal{P}$ & Technology with a resource availability capacity\\
    % $\mathcal{Q}'  $ & Set of technologies with \\
    % &\hspace{-0.2cm} resource resource availability capacity \\
    $\mathcal{T}^{\text{e}}$& Representative hours for power system\\
    $\mathfrak{R}$& Representative days\\
    $\mathfrak{T}^{\text{e}}_\tau$& Hours in the representative day $\tau$\\
    $\mathcal{S}^{\text{e}}_n$ &  Storage facility types \\
    $\mathcal{A}^{\text{g}}_n$ & Adjacent NG  nodes for node  $n$\\
    &\hspace{-0.2cm} between node $n$ and $m$\vspace{0.1cm}\\ 
    \hdashline \vspace{-0.3cm}\\
    $ \mathcal{N}^{\text{g}}, \mathcal{N}^s$ & NG and SVL nodes\\
     $\mathcal{T}^{\text{g}}$& Days of the planning year for NG system\\
    %$\mathcal{A}^{\text{e}}_k$ & Adjacent power system nodes for node  $k$\\
    $\mathcal{A}^s_k$ & Adjacent SVL facilities of node  $k$\\
    %$\mathcal{A}^{\text{g}}_j$ & All adjacent SVL facilities of node  $k\in \mathcal{N}^{\text{g}}$\\
$\mathcal{L}^{\text{g}}$ & Existing and candidate pipelines\\
    $\mathcal{L}^{\text{gExp}}_{k}$ & Existing and candidate pipelines starting from node $k$\\
    $\mathcal{L}^{\text{gImp}}_{k}$ & Existing and candidate pipelines ending at node $k$\\
    \bottomrule
    \end{tabular}
\end{table}

\begin{table}
\small
    \centering
   % \caption{Annualized cost parameters}
    \begin{tabular}{l l} 
          \multicolumn{2}{l}{{Annualized Cost Parameters}}\\
      \midrule
         $C^{\text{inv}}_{i}$ &  CAPEX of plants, [$\$$/plant] \\
        $C^{\text{dec}}_i$ &  Plant decommissioning cost, [$\$$/plant]\\
        $C^{\text{EnInv}}_{r}$ & Storage establishment energy-related cost, [$\$$/MWh]\\
       $C^{\text{pInv}}_{r}$ &  Storage establishment power-related cost, [$\$$/MW]\vspace{0.1cm}\\
     %$C_{\text{CO}_2}^{\text{inv}}$ & Levelized investment and FOM cost \\     & of CO$^2$ for pipeline [$\$$/ton/mile]\\
    %$C_{\text{CO}_2}^{\text{str}}$ & Levelized investment and FOM cost \\     &  of CO$^2$  storage [$\$$/ton]\\
     \hdashline \vspace{-0.3cm}\\
     
      ${C}^{\text{pipe}}_{\ell}$ &  Pipelines establishment cost, [$\$$/line] \\
       $C^{\text{strInv}}_{j}$ &  CAPEX of storage tanks at SVLs, [$\$$/MMBtu]\\
       $C^{\text{vprInv}}_{j}$ &  CAPEX of vapor. plants at SVLs, [$\$$/MMBtu]\\
     \bottomrule
    \end{tabular}
\end{table}

\begin{table}
\small
    \centering
    %\caption{Other cost parameters}
    \begin{tabular}{l l}
          \multicolumn{2}{l}{{Annual Costs}}\\
      \midrule
         $C^{\text{fix}}_{i}$ & FOM for plants, [$\$$] \\
        $C^{\text{EnFix}}_{r}$ & Energy-related FOM for storage, [\$/MWh]\\
        $C^{\text{pFix}}_{r}$ & Power-related FOM for storage, [\$/MW] \vspace{0.1cm}\\  \hdashline \vspace{-0.3cm}\\  
      $C^{\text{strFix}}_{j}$ & FOM for storage tanks, [\$/MMBtu]\\
      $C^{\text{vprFix}}_{j}$ & FOM for vaporization plants, [\$/MMBtu]\\
     %$C^{\text{liqFix}}_{j}$ & FOM for liquefaction plants, [$\$$/MMBTu]\\
     \bottomrule
    \end{tabular}
\end{table}

\begin{table}
\small
    \centering
    %\caption{Other cost parameters}
    \begin{tabular}{l l}
          \multicolumn{2}{l}{{Other Cost Parameters}}\\
      \midrule
        $C^{\text{var}}_{i}$ & VOM for  plants, [$\$$/MWh]\\
    %$C^{\text{startUp}}_i$ & Start-up cost for plants, [$\$$]\\
     $C^{\text{eShed}}$ & Unsatisfied power demand cost, [$\$$/MWh] \\
      $C^{\text{fuel}}_i$ & Fuel price for plants, [$\$$/MMBtu]\vspace{0.1cm}\\  \hdashline \vspace{-0.3cm}\\
     %$C_{\text{CO}_2}^{\text{var}}$ & VOM cost of CO$^2$ pipelines, [$\$$/ton]\\
     $C^{\text{ng}}$ & Fuel price for NG, [$\$$/MMBtu]\\
     $C^{\text{rng}}$ & Price of RNG, [$\$$/MMBtu]\\
     $C^{\text{gShed}}$ & Unsatisfied NG demand cost [$\$$/MMBtu]\\
     \bottomrule
    \end{tabular}
\end{table}

\begin{table}
\small
    \centering
    %\caption{Parameters for Power System}
    \begin{tabular}{l l}
          \multicolumn{2}{l}{{Other Parameters for the Power System}}\\
      \midrule
         $\rho_{nti}$ & Capacity factor \\
         $D^{\text{e}}_{nt}$ & Power demand, [MWh]\\
       $h_i$ & Heat rate, [MMBtu/MWh]\\
        $\eta_i$ & Carbon capture rate, [\%]\\
        %$\eta^{\text{g}}$ & Emission rate of natural gas, [ton CO$^2$/MMBtu]\\
       $U^{\text{prod}}_{i}$ & Nameplate capacity, [MW]\\
        $L^{\text{prod}}_{i}$ & Minimum stable output, [$\%$]\\
        $U^{\text{ramp}}_{i}$ & Ramping limit, [$\%$]\\
        $\gamma^{\text{eCh}}_r,\gamma^{\text{eDis}}_r$ & Charge/discharge rate for storage\\\
        ${I}^{\text{num}}_{ni}$ & Initial number of plants\\
        $U^{\text{e}}_{\text{emis}}$ & Baseline emission of CO$_2$ in 1990 \\
        &from generation consumption, [ton]\\
       %$U^{\text{CCS}}$ & Total annual carbon storage capacity, [ton]\\
        $L^{\text{RPS}}$ & Renewable Portfolio Standard (RPS) value\\        
        $\zeta$ & Emission reduction goal\\
        $w_t$ & Number of periods represented by period $t$\\
        $\phi^{\text{e}}_t$& Mapping of representative period $t$ to its\\
        & original period in the time series\\
        \bottomrule
    \end{tabular}
\end{table}

\begin{table}[ht]
\small
    \centering
    %\caption{Other parameters for NG Systems}
    \begin{tabular}{l l}
          \multicolumn{2}{l}{{Other Parameters for the NG System}}\\
      \midrule
         $D^{\text{g}}_{k\tau}$ &  NG demand, [MMBtu]\\
        $\eta^{\text{g}}$ & Emission factor for NG [ton CO$_2$/MMBtu]\\
        $U^{\text{inj}}_k$ & Upper bound for NG supply, [MMBtu]\\
        $\gamma^{\text{liqCh}}_j$ & Charge efficiency of liquefaction plant\\
        $\gamma^{\text{vprDis}}_j$ & Discharge efficiency of vaporization plant\\
        $\beta$ & Boil-off gas coefficient\\
        $I^{\text{pipe}}_{\ell}$ & Initial capacity for pipeline $\ell$, [MMBtu]\\
        $U^{\text{pipe}}_{\ell}$ & Upper bound capacity for pipeline $\ell$, [MMBtu]\\
        $\mathcal{I}^{\text{pipe}}_{\ell}$ & 1, if the pipeline $\ell$ exists; 0, otherwise\\
        $I^{\text{gStr}}_{j}$ & Initial storage capacity, [MMBtu]\\
        $I^{\text{vpr}}_{j}$ & Initial vaporization capacity, [MMBtu/d]\\
        $I^{\text{liq}}_{j}$ & Initial liquefaction capacity, [MMBtu/d]\\
        $I^{\text{store}}_{kj}$ & Initial capacity of storage facility\\
        %$U^{\text{rng}}$ & Total available RNG, [MMBtu]\\
        $U^{\text{g}}_{\text{emis}}$ & Baseline emission of CO$_2$ in 1990\\
        &from non-generation consumption, [ton]\\
        $\Omega_\tau$ & Set of days represented by day $\tau \in \mathfrak{R}$\\
        \bottomrule
    \end{tabular}
\end{table}

\begin{table}[ht]
\small
    \centering
    %\caption{Investment Decision Variables}
    \begin{tabular}{l l}
         \multicolumn{2}{l}{{Investment Decision Variables}}\\
      \midrule
         $x^{\text{op}}_{ni}\in \mathbb{Z}^+$ & Number of available plants\\         
         $x^{\text{est}}_{ni}\in \mathbb{Z}^+$ & Number of new plants established \\
         $x^{\text{dec}}_{ni}\in \mathbb{Z}^+$ & Number decommissioned plants \\
         $y^{\text{eCD}}_{nr}\in \mathbb{R}^+$& Charge/discharge capacity of storage battery\\
         $y^{\text{eLev}}_{nr}\in \mathbb{R}^+$& Battery storage level\\
         $z_{\ell}^{\text{g}}\in \mathbb{B}$& 1, if pipeline $\ell$ is built; 0, otherwise\\
         \bottomrule
    \end{tabular}
\end{table}

\begin{table}[ht]
\small
    \centering
    %\caption{Other Decision Variables for Power System}
    \begin{tabular}{l l}
 \multicolumn{2}{l}{{Other Decision Variables for Power System}}\\
      \midrule 
$p_{nti}\in \mathbb{R}^+$& Generation rate, [MW]\\
         %$x_{nti}\in \mathbb{R}^+$ & Number of committed plants\\
         %$x^{\text{down}}_{nti}\in \mathbb{R}^+$ & Number of plants shut-down\\
         %$x^{\text{up}}_{nti}\in \mathbb{R}^+$ & Number of plants started up\\
         $f^{\text{e}}_{\ell t} \in {\mathbb{R}}$& Flow rates, [MW]\\
        %$\theta_{nt}\in \mathbb{R}$ & Phase angle\\
        $s^{\text{eCh}}_{ntr},s^{\text{eDis}}_{ntr} \in \mathbb{R}^+$& Storage charged/discharged, [MW] \\
        $s^{\text{eLev}}_{ntr} \in \mathbb{R}^+$& Storage level, [MWh] \\
         $a^{\text{e}}_{nt}\in \mathbb{R}^+$ & Amount of load shedding, [MWh]\\
         $\mathcal{E}^{\text{e}}$ & Total emission from power system\\
        \bottomrule
    \end{tabular}
\end{table}

\begin{table}[ht]
\small
    \centering
    %\caption{Other Decision Variables for NG System (all in MMBtu)}
    \begin{tabular}{l l}
 \multicolumn{2}{l}{Other Decision Variables for NG System (all in MMBtu)}\\
      \midrule  
        $f^{\text{g}}_{\ell\tau} \in {\mathbb{R}}^+$& Flow rates between NG nodes\\
         $g_{k\tau }\in \mathbb{R}^+$& NG supply (injection)\\
        %$s^{\text{gLev}}_{k\tau j} \in \mathbb{R}^+$& Storage levels \\
         $a^{\text{g}}_{k\tau}\in \mathbb{R}^+$ & Amount of load shedding\\
        $a^{\text{rng}}_{k\tau}\in \mathbb{R}^+$ & Amount of RNG consumption\\
        $\mathcal{E}^{\text{g}}$ & Total emission from NG system\\
        \bottomrule
    \end{tabular}
\end{table}

\subsection*{Power System Model}
\noindent\textbf{Objective Function:}
\begin{subequations}
\label{elec-obj}
\begin{align}
      \min & \sum_{n \in \mathcal{N}^{\text{e}}} \sum_{i \in \mathcal{P}}  (C^{\text{inv}}_i x^{\text{est}}_{ni}+C^{\text{fix}}_{i} x^{\text{op}}_{ni} + \sum_{r \in \mathcal{S}^{\text{e}}_n}(C^{\text{pInv}}_r+C^{\text{pFix}}) y^{\text{eCD}}_{nr})+\notag\\
       &\sum_{n \in \mathcal{N}^{\text{e}}} \sum_{r \in \mathcal{S}^{\text{e}}_n}(C^{\text{EnInv}}_r+C^{\text{EnFix}}) y^{\text{eLev}}_{nr} +  \label{elec-obj-1}\\
        & \sum_{n \mathcal{N}^{\text{e}}} \sum_{i \in \mathcal{P}}  C^{\text{dec}}_i x^{\text{dec}}_{ni} \label{elec-obj-2}\\
    &  \sum_{n \in \mathcal{N}^{\text{e}}}\sum_{i \in \mathcal{P}} \sum_{t \in \mathcal{T}^{\text{e}}}w_t p_{nti}C^{\text{var}}_{i} + \label{elec-obj-3}\\  
         &  \sum_{n \in \mathcal{N}^{\text{e}}}\sum_{i \in \mathcal{P}} \sum_{t \in \mathcal{T}^{\text{e}}}w_t p_{nti} (C^{\text{fuel}}_{i} h_i) + \label{elec-obj-7}\\
    &\sum_{n \in \mathcal{N}^{\text{e}}} \sum_{t \in \mathcal{T}^{\text{e}}}w_t C^{\text{eShed}}_n a^{\text{e}}_{nt}+
    \label{elec-obj-8}
\end{align}
\end{subequations}

The objective function~\eqref{elec-obj} minimizes the  total investment and operating costs incurred in power system. The first term~\eqref{elec-obj-1} is the investment and fixed operation and maintenance (FOM) costs for generation and storage. The term~\eqref{elec-obj-2} captures the cost of plant retirement or decommissioning. The variable operating and maintenance  (VOM) are represented by term~\eqref{elec-obj-3}. The cost of fuel consumption for non-gas-fired power plants (i.e., nuclear plant) are ensured by term~\eqref{elec-obj-7}.  The term term~\eqref{elec-obj-8} penalizes the load shedding in the power system which can occur due to unsatisfied demand. 

\noindent\textbf{Investment:} For every $n\in \mathcal{N}^{\text{e}},i\in \mathcal{P}$
\begin{subequations}
\begin{align}
      & x^{\text{op}}_{ni} = I^{\text{num}}_{ni}-x^{\text{dec}}_{ni}+x^{\text{est}}_{ni} &  \label{elec-c1}\\
%   &x_{nti} - x_{n,t-1,i} = x^{\text{up}}_{nti}-x^{\text{down}}_{nti}& t\in \mathcal{T}^{\text{e}} \label{elec-c2}\\
%     &x_{nti} \leq x^{\text{op}}_{ni}&t\in \mathcal{T}^{\text{e}} \label{elec-c3}
\end{align}
\end{subequations}
constraints~\eqref{elec-c1} specify the number of operating plants.

\noindent\textbf{Generation, Ramping, and Load Shedding:} For every $n \in \mathcal{N}^{\text{e}},t\in \mathcal{T}^{\text{e}}$
\begin{subequations}
\begin{align}
    %% Generation limit
    &L^{\text{prod}}_{i} U^{\text{prod}}_{i} x^{\text{op}}_{ni} \leq  p_{nti} \leq   U^{\text{prod}}_{i}x^{\text{op}}_{ni}&\hspace{-2cm} i\in \mathcal{H} %[\beta_{nti}] 
    \label{elec-c4}\\
    % ramping 
    &|p_{nti} -p_{n,(t-1),i}| \leq   U^{\text{ramp}}_{i} U^{\text{prod}}_{i} x^{\text{op}}_{ni}+\notag\\
    &\max(L^{\text{prod}}_{i}, U^{\text{ramp}}_{i})U^{\text{prod}}_{i}x^{\text{op}}_{ni}   & \hspace{-2cm} i\in \mathcal{H} %[\gamma_{nti}] 
    \label{elec-c5}\\
     &p_{nti} \leq \rho_{nti}U^{\text{prod}}_{i} x^{\text{op}}_{ni} &\hspace{-2cm} i \in \mathcal{R}  \label{elec-c6}\\
     & a^{\text{e}}_{nt}\leq D^{\text{e}}_{n\phi^{\text{e}}_t}& \label{elec-c7} 
\end{align}
\end{subequations}
the generation limits are imposed in constraints~\eqref{elec-c4}. 
Constraints~\eqref{elec-c5} are the ramping constraints that limit the  generation difference of thermal units in any consecutive time periods to a ramping limit in the right-hand-side of the equation. The generation pattern of VREs is determined by their hourly profile in the form of capacity factor; constraints~\eqref{elec-c6} limit the generation of VRE to hourly capacity factor (i.e. $\rho_{nti}$) of maximum available capacity (i.e. $U^{\text{prod}}_{i}x^{\text{op}}_{ni}$). Constraints~\eqref{elec-c7} state that the load shedding amount can not exceed demand.

\noindent\textbf{Power Balance Constraints:}
For every  $t\in \mathcal{T}^{\text{e}}$

\begin{subequations}
\begin{align}
      % Power balance
        &\sum_{n \in \mathcal{N}^e} ( \sum_{i \in \mathcal{P}}p_{nti} +
        \sum_{r\in \mathcal{S}^{\text{e}}_n} (s^{\text{eDis}}_{ntr}-s^{\text{eCh}}_{ntr})+a^{\text{e}}_{nt})=\sum_{n \in \mathcal{N}^e} D^{\text{e}}_{n \phi^{\text{e}}_t}
    & \hspace{-2cm} %[\theta_{nt}]
    \label{elec-c8}
\end{align}
\end{subequations}
constraints~\eqref{elec-c8} ensure the balance of supply and demand in the system. In particular, it ensures that over all nodes, the generation, the net storage power, and the power load shedding is equal to the demand of all nodes. 

\noindent\textbf{Storage Constraints:} For every $n\in \mathcal{N}^{\text{e}}, t\in \mathcal{T}^{\text{e}}, r\in \mathcal{S}^{\text{e}}_n$
\begin{subequations}
\begin{align}
 &s^{\text{eLev}}_{ntr} = s^{\text{eLev}}_{n,t-1,r}+\gamma^{\text{eCh}}_r s^{\text{eCh}}_{ntr}-\frac{s^{\text{eDis}}_{ntr}}{\gamma^{\text{eDis}}_r}&  \label{elec-c15}\\
    &s^{\text{eDis}}_{ntr}\leq  y^{\text{eCD}}_{nr}, s^{\text{eCh}}_{ntr}\leq y^{\text{eCD}}_{nr}  &  \label{elec-c16}\\
    & s^{\text{eLev}}_{ntr}\leq y^{\text{eLev}}_{nr} &  \label{elec-c17}\\
    &s^{\text{eLev}}_{n,t_1,r}=s^{\text{eLev}}_{n,t_{24},r}  &\hspace{-1cm} t_1, t_{24}\in \mathfrak{T}^{\text{e}}_\tau, \tau \in \mathfrak{R}  \label{elec-c18-2}
    % yCD >= Sdis+Sch
\end{align}
\end{subequations}
Constraints~\eqref{elec-c15} model battery storage dynamics. The charge/discharge limits are imposed in ~\eqref{elec-c16}, and constraints~\eqref{elec-c17} limits the storage level. Note that as for other similar studies \cite{SepulvedaEtal2021,LiEtal2022}, we do not account for storage capacity degradation. Representative days are not necessarily consecutive, therefore the formulation should account for the carryover storage level between representative days. Li et al. \cite{LiEtal2022} enforce the beginning and ending storage levels of each representative days to 50\% of the maximum storage level. In constraints~\eqref{elec-c18-2}, we use a similar technique, yet more flexible, as we assume that beginning (i.e., $t_1$) and ending (i.e., $t_{24}$) storage levels are the same for any representative day.

\noindent\textbf{Renewable Portfolio Standards (RPS):} 
\begin{subequations}
\begin{align}
   & \sum_{n\in \mathcal{N}^{\text{e}}}\sum_{t\in \mathcal{T}^{\text{e}}}\sum_{i\in \mathcal{R}}  p_{nti} \geq L^{\text{RPS}}\sum_{n\in \mathcal{N}^{\text{e}}}\sum_{t\in \mathcal{T}^{\text{e}}} D^{\text{e}}_{n\phi^{\text{e}}_t}  %[\omega]
   \label{elec-c18}
\end{align}
\end{subequations}
The formulation requires the model to procure a certain share of the total demand from renewable energy sources. The share of renewable energy sources which is known as Renewable Portfolio Share (RPS) is imposed by constraint~\eqref{elec-c18}.

\subsection{NG System Model}
%This section presents the formulation for NG system. 
\noindent\textbf{Objective Function:}
\begin{subequations}\label{ng-obj}
\begin{align}
    \min \ & \sum_{l\in \mathcal{L}^{\text{g}}} C^{\text{pipe}}_{\ell} z_{\ell}^{\text{g}} +  \label{ng-obj-1}\\
    & \sum_{k \in \mathcal{N}^{\text{g}}} \sum_{\tau \in \mathcal{T}^{\text{g}}}  C^{\text{ng}} g_{k\tau}+\label{ng-obj-2}\\
    & \sum_{k \in \mathcal{N}^{\text{g}}} \sum_{t \in \mathcal{T}^{\text{g}}} (C^{\text{rng}}a^{\text{rng}}_{k\tau } +C^{\text{gShed}} a^{\text{ng}}_{k\tau })\label{ng-obj-5}
\end{align}
\end{subequations}

The objective function~\eqref{ng-obj} minimizes the total investment and operating costs incurred in the NG system. The first term~\eqref{ng-obj-1} is the investment cost for establishing new pipelines. The second term~\eqref{ng-obj-2} is the cost of procuring NG from various sources to the system. For example, New England procures its NG from Canada, and its adjacent states such as New York. The last term~\eqref{ng-obj-5} captures the cost of using RNG and NG load shedding.  

\noindent\textbf{NG Balance Constraint:}  For every $k\in \mathcal{N}^{\text{g}}, \tau \in \mathcal{T}^{\text{g}} $
\begin{subequations}
\begin{align}
   &g_{k\tau} -\sum_{l \in \mathcal{L}^{\text{gExp}}_{k}} f^{\text{g}}_{\ell\tau}+\sum_{l \in \mathcal{L}^{\text{gImp}}_{k}} f^{\text{g}}_{\ell\tau}-\sum_{n\in \mathcal{A}^{\text{e}}_k} f^{\text{ge}}_{kn\tau } + a^{\text{rng}}_{k\tau}+a^{\text{g}}_{k\tau}=D^{\text{g}}_{k\tau} &  \label{ng-c1}
\end{align}
\end{subequations}
constraints~\eqref{ng-c1} state that for each node and period, the imported NG (i.e., injection), flow to other NG nodes, flow to power nodes, satisfied load by RNG, and unsatisfied NG load should add up to demand. The flow in pipelines are modeled unidirectional as it is typical  for most long-distance transmission pipelines involving booster compressor stations \cite{VonWaldEtal2022}. We are ignoring electricity consumption associated with booster compression stations along the NG pipeline network.  Note that there is no load shedding in NG system as we assume  RNG availability for any quantity.

% \noindent\textbf{Representative Days:}
% \begin{subequations}
% \begin{align}
%    &f^{\text{ge}}_{kn\tau_1 } = f^{\text{ge}}_{kn\tau_2 }& \tau_1, \tau_2 \in \Omega_\tau, \tau\in \mathfrak{R} \label{ng-c2-1}
% \end{align}
% \end{subequations}
% The constraint~\eqref{ng-c2-1} captures the impact of representative days on the gas system. It ensures that gas consumption by the power system for all the days in the same cluster is the same. 

\noindent\textbf{Gas and RNG Supply Constraints:}
For every $k\in \mathcal{N}^{\text{g}}, \tau \in \mathcal{T}^{\text{g}}$
\begin{subequations}
\begin{align}
   & L^{\text{inj}}_k \leq g_{k\tau}\leq U^{\text{inj}}_k & \label{ng-c2}\\
   & a^{\text{rng}}_{k\tau}+a^{\text{g}}_{k\tau}\leq D^{\text{g}}_{k\tau} & \label{ng-c3}\\
   &a^{\text{rng}}_{k\tau}\leq U^{\text{inj}}_k M&\label{ng-c3-2}
\end{align}
\end{subequations}
import limits are imposed in constraints~\eqref{ng-c2}. The consumption of RNG plus the load shedding is limit by constraints~\eqref{ng-c3} to the NG load. The alternative fuel RNG can only be imported from injection points as specified by constraints~\ref{ng-c3-2}.

\noindent\textbf{Flow Constraints:}
For every $\ell \in \mathcal{L}^{\text{g}}, \tau \in \mathcal{T}^{\text{g}}, j\in \mathcal{N}^s$
\begin{subequations}
\begin{align}
       & f^{\text{g}}_{\ell \tau } \leq I^{\text{pipe}}_{\ell}& \text{if } \mathcal{I}^{\text{pipe}}_{\ell}=1 \label{ng-c5}\\
    & f^{\text{g}}_{\ell \tau } \leq U^{\text{pipe}}_{\ell} z^{\text{g}}_{\ell}& \text{if } \mathcal{I}^{\text{pipe}}_{\ell}=0  \label{ng-c6}\\
\end{align}
\end{subequations}
constraints~\eqref{ng-c5} and \eqref{ng-c6}  limit the flow between NG nodes for existing and candidate pipelines, respectively. 

\subsection{Coupling Constraints}
The following constraints are coupling constraints that relate decisions of the two systems.
\begin{subequations}
\begin{align}
    %&  f^{\text{ge}}_{k n\tau } =   \sum_{t \in \mathfrak{T}^{\text{e}}_\tau} \sum_{i \in \mathcal{G}} h_i p_{nti} &\hspace{-2cm} k\in \mathcal{N}^{\text{g}},n\in \mathcal{A}^{\text{e}}_k, \tau \in \mathfrak{R}\label{coup-1}\\
    & \sum_{k \in \mathcal{A}^{\text{e}}_n} f^{\text{ge}}_{k n\tau } =   \sum_{t \in \mathfrak{T}^{\text{e}}_\tau} \sum_{i \in \mathcal{G}} h_i p_{nti} &\hspace{-2cm} n\in \mathcal{N}^{\text{e}}, \tau \in \mathfrak{R}\label{coup-1}\\
 &\mathcal{E}^{\text{e}} = \sum_{n\in \mathcal{N}^{\text{e}}}\sum_{t\in \mathcal{T}^e}\sum_{i \in \mathcal{G}} w_t(1-\eta_i)\eta^{\text{g}} h_i p_{nti}&\notag \\
 &\mathcal{E}^{\text{g}} =\sum_{k \in \mathcal{N}^g}\sum_{\tau \in \mathcal{T}^g} \eta^{\text{g}}( D^{\text{g}}_{k\tau}-  a^{\text{rng}}_{k\tau}-a^{\text{g}}_{k\tau})\notag \\
 &  \mathcal{E}^{\text{e}}+\mathcal{E}^{\text{g}} \leq (1-\zeta) (U^{\text{e}}_{\text{emis}} +U^{\text{g}}_{\text{emis}}) &\label{coup-2}
\end{align}
\end{subequations}
The first coupling constraints~\eqref{coup-1} captures the flow of NG to the power network for each node and at each time period. 
The variable $\mathcal{E}^{\text{e}}$ accounts for the emission due to the consumption of NG in the power system.  The variable  $\mathcal{E}^{\text{g}}$ computes the emission from NG system by subtracting the demand from RNG consumption. The second coupling constraint~\eqref{coup-2} ensures that the net CO$_2$ emissions associated with electric-NG system is below the specified threshold value, which is defined based on reduction relative to some baseline emissions.Since the model cannot track whether RNG is used to meet non-power NG demand or for power generation, the constraint  ~\eqref{coup-2}  first computes gross emissions from all NG use presuming it is all fossil and then subtracts emissions benefits from using RNG. Here we treat RNG as a carbon-neutral fuel source, and thus the combustion emissions associated with its end-use are equal to the emissions captured during its production. Recent life cycle analysis studies suggest that depending on the feedstock RNG could have negative to slightly positive life cycle GHG emissions \cite{LeeEtal2021}. 

The first term is the emission due to non-generational NG consumption (i.e., NG consumption in the NG system such as space heating, industry use, and transportation) and the second term captures the emission from gas-fired power plants. Alternatively, the emission constraints can only be applied to the power system as in \cite{SepulvedaEtal2021} or separately applied to each system as in \cite{VonWaldEtal2022}. 

\end{document}